\documentclass{article}

\usepackage{pgfplots}
\usepackage[table]{xcolor}
\usepackage{plain}
\usepackage{graphicx} % Required for inserting images
\usepackage{adjustbox}
\usepackage{tabularx}
\usepackage{standalone}
\usepackage{tikz}
\usetikzlibrary{calc}
\usetikzlibrary{fit}
\usetikzlibrary{backgrounds}
\usetikzlibrary{external}
\usepackage{multicol}
\usepackage{amsmath, amssymb}
\usepackage{ulem}
\usepackage{subcaption}
\usepackage[utf8]{inputenc} % se compili con pdfLaTeX
\usepackage[T1]{fontenc}
\usepackage{pmboxdraw}
\usepackage{wrapfig}

\usepackage{booktabs}
\usepackage[most]{tcolorbox}

\definecolor{mycolor}{HTML}{2A4759}
\definecolor{Tron primary}{HTML}{193f4a}
\definecolor{Tron tern}{HTML}{2f8ca3}
\definecolor{Jungle green}{HTML}{174312}

\definecolor{Megatron light grey}{HTML}{a5a7aa}
\definecolor{Megatron grey}{HTML}{A8A9AD}
\definecolor{Megatron dark grey}{HTML}{71706E}
\definecolor{Decepticon purple}{HTML}{8712B6}

\usepackage{lipsum}
\usepackage{amsfonts}
\usepackage{float}
\title{A Multi-Label Temporal Convolutional Framework for Transcription
Factor Binding Characterization}
\author{P. Demurtas, F. Zanchetta, G. Perini, R. Fioresi}
\usepackage{pgfplots}

\usepgfplotslibrary{colormaps} % colormap 'viridis', 'plasma', 'RdBu', ecc.
\pgfplotsset{compat=newest}

\begin{document}

    % ========= MACRO ROBUSTE (niente \edef) =========

% Valori in [0,1] su colormap sequenziale (es. viridis/plasma)
% Uso: \CellColor01{val∈[0,1]}{nome-colormap}{contenuto}
\DeclareRobustCommand{\CellColor01}[3]{%
    \begingroup
    % clamp in [0,1] usando pgfmath
    \pgfmathsetmacro{\t}{min(1,max(0,#1))}%
    % \pgfplotscolormapaccess restituisce {R,G,B} in \pgfmathresult
    \pgfplotscolormapaccess[0:1]{\t}{#2}%
    % applica il colore alla cella (senza \edef)
    \edef\CellColor@call{\noexpand\cellcolor[rgb]{\pgfmathresult}}%
    \CellColor@call\ignorespaces#3%
    % \cellcolor[rgb]{\pgfmathresult}#3%
    %\cellcolor[rgb]{0.129411,0.568621,0.549009} PIPPO
    \endgroup
}

% Celle "media ± sd" (valori già in [0,1])
% Uso: \MetricCell01{media}{sd}{nome-colormap}
\DeclareRobustCommand{\MetricCell01}[3]{%
    \CellColor01{#1}{#3}{$#1\pm#2$}%
}

% Δ in [-R,+R] su colormap divergente centrata a 0 (default R=1)
% Uso: \CellColorSigned{val}{R}{nome-colormap}{contenuto}
\DeclareRobustCommand{\CellColorSigned}[4]{%
    \begingroup
    % mappa [-R,R] -> [0,1] e clamp
    \pgfmathsetmacro{\t}{(\numexpr 0\relax + #1 + #2)/(2*#2)}%
    \pgfmathsetmacro{\t}{min(1,max(0,\t))}%
    % colormap -> {R,G,B} in \pgfmathresult
    \pgfplotscolormapaccess[0:1]{\t}{#3}%
    % applica colore
    \edef\CellColor@call{\noexpand\cellcolor[rgb]{\pgfmathresult}}%
    \CellColor@call\ignorespaces#4%
    \endgroup
}

\pgfplotscreatecolormap{viridis}{
    rgb255(0cm)=(68,1,84);
    rgb255(0.25cm)=(59,82,139);
    rgb255(0.50cm)=(33,145,140);
    rgb255(0.75cm)=(94,201,97);
    rgb255(1cm)=(253,231,37)
}
\pgfplotscreatecolormap{plasma}{
    rgb255(0cm)=(13,8,135);
    rgb255(0.25cm)=(126,3,167);
    rgb255(0.50cm)=(203,71,119);
    rgb255(0.75cm)=(248,149,64);
    rgb255(1cm)=(240,249,33)
}
% Divergente "RdBu reversed" SENZA spazi nel nome (usa questo nome!)
\pgfplotscreatecolormap{RdBureversed}{
    rgb255(0cm)=(33,102,172);
    rgb255(0.50cm)=(247,247,247);
    rgb255(1cm)=(178,24,43)
}

% --- Scegli le colormap globali (coerenti su tutte le tabelle) ---
\newcommand{\TCNmap}{viridis}
\newcommand{\RNNmap}{viridis}%{plasma}
\newcommand{\DIVERGmap}{RdBu} % useremo reversed per avere blu=neg, rosso=pos
% \pgfplotsset{colormap/\DIVERGmap reversed}

    \newcommand{\gzero}{D-5TF-3CL}
\newcommand{\gone}{D-7TF-4CL}

{\bf
\maketitle}
%\tableofcontents

    \section*{Abstract}
    Transcription factors (TFs) regulate gene expression through complex and cooperative
    mechanisms. While many TFs act together, the logic underlying TFs binding and their
    interactions is not fully understood yet. Most current approaches for TF binding site prediction
    focus on individual TFs and binary classification tasks, without a full analysis of
    the possible interactions among various TFs.
    In this paper
    we investigate DNA TF binding site recognition as a multi-label classification problem,
    achieving reliable predictions for multiple TFs on DNA sequences
    retrieved in public repositories. Our deep learning models are based on
    Temporal Convolutional Networks (TCNs), which are
    able to predict multiple TF binding profiles,
    capturing correlations among TFs and their cooperative
    regulatory mechanisms.
Our results suggest that multi-label learning
leading to reliable predictive performances can reveal biologically meaningful motifs
and co-binding patterns consistent with known TF interactions,
while also suggesting novel relationships and cooperation among TFs.
%Our work highlights the potential of multi-label deep learning frameworks
%combined with explainability to advance the understanding of transcriptional
%regulation and TFs cooperation directly from genomic sequence.
%    \listoftables
%    \listoffigures
    \section{Introduction}
%{Biological motivation}

    Transcription factors (TFs) rarely act alone, but they
    frequently operate through cooperative mechanisms,
    forming complexes such as homo/hetero dimers, where distinct combinations of TFs
can elicit different regulatory effects~~~
\cite{latchman1997transcription,spitz2012transcription,Xie2025DNAguidedTF}.
    In prokaryotes, individual TFs typically recognize relatively long DNA motifs,
    which are often sufficient to uniquely identify their target genes.
    In contrast, TFs in organisms with larger and more complex genomes bind shorter
    DNA sequences, which are insufficient to define unique genomic
    locations on their own. Moreover, the development and maintenance of multicellular organisms
    require the emergence of intricate molecular systems capable of implementing
    combinatorial regulatory logic~\cite{morgunova2017structural}.
    To overcome these challenges, eukaryotic organisms have evolved mechanisms for cooperative
    DNA recognition involving multiple TFs,
%Such cooperation comes
    through direct  and indirect protein-protein interactions~\cite{reiter2017direct},
    indirect interactions mediated by chromatin architecture, or through co-binding
    to adjacent or partially overlapping DNA motifs~\cite{alon2007network}.
    Each interaction mechanism confers distinct regulatory properties to the
    resulting TF complex~\cite{morgunova2017structural}.

A representative example of TF cooperation is the formation of functional heterodimers
\cite{Xie2025DNAguidedTF}.
    Several eukaryotic TFs are unable to bind DNA as monomers and instead require physical
interaction with another TF, often a member of the same family,
%non so quanto sia biologicamente corretta,
to form a functional dimer capable of recognizing specific DNA sequences.

%    \medskip\hrule\medskip
%    \textcolor{red}{PERINI: Check above statements.} PERINI AGREES
%    \medskip\hrule\medskip

    \begin{wrapfigure}{l}{0.5\textwidth}
        \centering
        \includegraphics[width=1\linewidth]{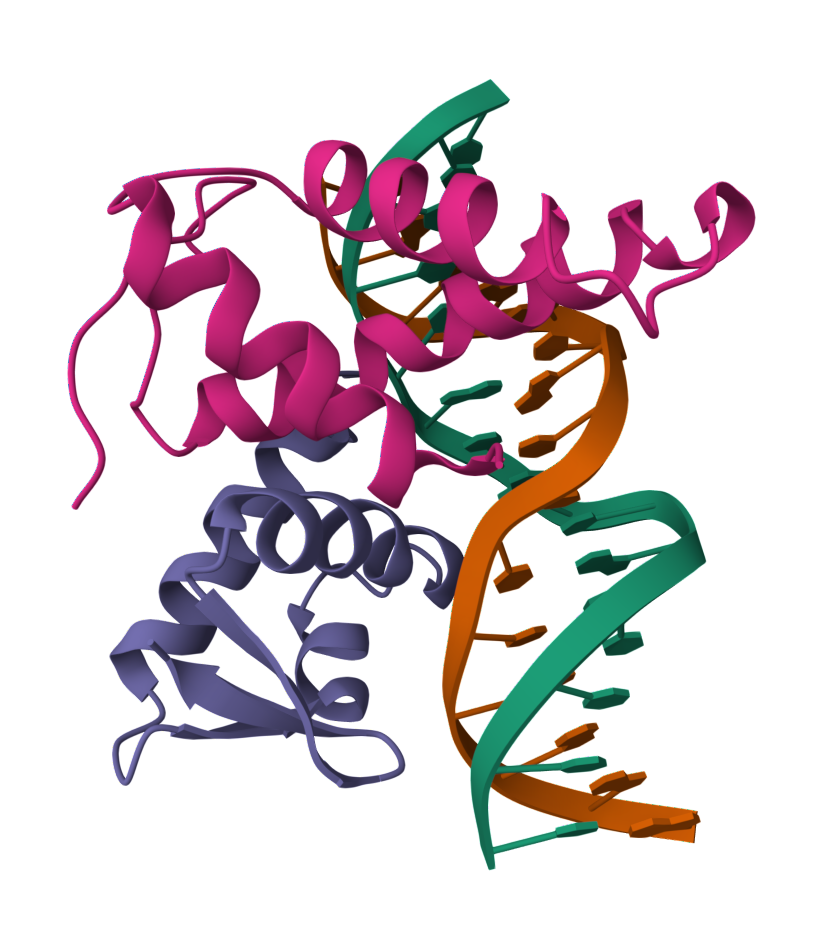}
        \caption{E2F4-DP2-DNA complex \cite{zheng1999structural}}
        \label{fig:E2F4-DNA}
    \end{wrapfigure}

    A classical example of such a mechanism is the MYC/MAX heterodimer, %a powerful oncogene,
    which plays a central role in transcriptional regulation
%, acting on thousands of genes, depending on the type of tissues and tumours
    \cite{amati1992myc,ahmadi2021myc}.
    While homo/hetero dimerization represent common forms of TF cooperation,
    they are only a subset of a much broader spectrum of regulatory complexes,
%Many TF assemblies participate in highly interconnected gene regulatory networks
whose combinatorial logic and mechanisms of action remain largely unexplored
\cite{buchler2003combinatorial}.
%TF complexes are therefore key components and effectors of gene
%regulatory circuits that ensure proper cellular function.
%Despite their fundamental importance, many of their combinatorial and logical
%properties remain unknown, representing a major frontier in molecular biology.
%\subsection{Computational motivation}

In this paper we plan to investigate the question of multiple TFs DNA-binding prediction
as a multi-label classification  problem {via deep learning}.
%\footnote{Multi-label classification involves predicting zero or more non-exclusive class labels for each instance.} offers a natural computational framework to investigate TF interactions directly from sequence data.
    This perspective will enable the simultaneous prediction of binding events
    for multiple TFs and provides an opportunity to explore correlations among
    TFs that reflects the cooperative or combinatorial regulatory mechanisms
     %   {\color{red}
        as in Fig. \ref{fig:E2F4-DNA}, obtained via
        a costly procedure, but going beyond.
        Our techniques can indeed refine and guide investigation
        complementing expensive lab protocols .
    % [REF 44] Pietro - Io chiederei se possibile conferma a Perini o ad un biologo in generale ( anche per la ref alle tecniche)
    %    }

To date, deep learning approaches for DNA-binding recognition have %predominantly
focused on binary classification tasks, where the goal is to predict whether a given
genomic region is bound by a single TF~\cite{han2019,zhang2022novel, fioresi2022deep}.

    \begin{figure}[H]
        \centering
        \includegraphics[width=0.75\textwidth]{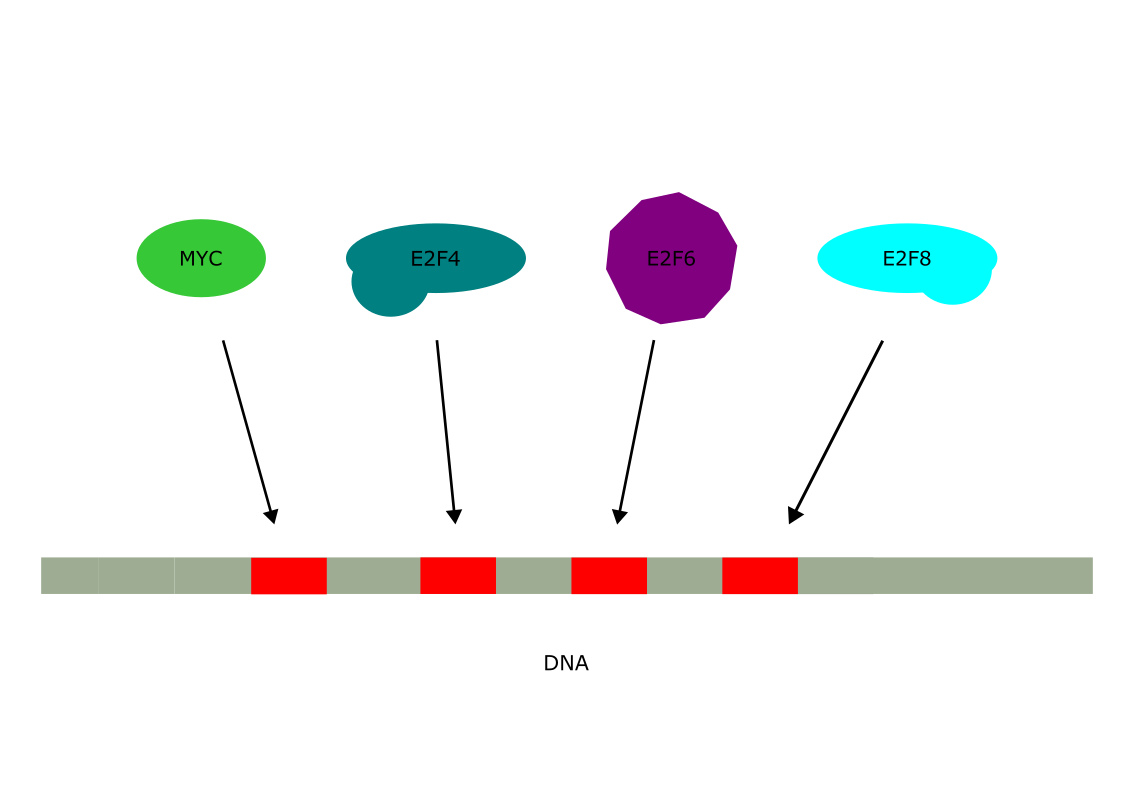}
        \caption{Multiple TFs binding to DNA}
        \label{fig:multi_tf}
    \end{figure}

    %In our previous work~\cite{fioresi2022deep},
    %we investigated the ability of deep learning
%models to learn TF-specific MYC consensus sequences.
In the present work, we take advantage of novel deep learning architectures, especially
designed for sequential data modelling as Temporal Convolutional Networks (TCNs)
for multi-label modelling TF binding site predictions, see Figure~\ref{fig:multi_tf}.
 TCNs exhibit a decisive advantage with respect to
traditional recurrent architectures, such as Recurrent Neural Networks (RNNs),
which have been widely applied to sequence modelling tasks,
including biological sequence analysis.
RNNs however suffer from well-known limitations, including vanishing and
exploding gradients~\cite{hochreiter1997long},
limited parallelizability, and difficulties in capturing long-range dependencies.
Also, more recently, attention-based models, most notably Transformers
\cite{vaswani2017attention}, have achieved state-of-the-art performance
across a wide range of sequence and language modelling
tasks~\cite{bengio2000neural,Devlin2019BERTPO}.
Their ability to model global dependencies and their scalability have made
them highly successful in many domains.
Nevertheless, attention-based architectures~\cite{brauwers2021general}
come with significant drawbacks, including substantial data requirements,
high computational costs during training, and limited interpretability.
    These issues are particularly problematic in biological applications,
    where data are often noisy or scarce and model transparency is essential.
Temporal Convolutional Networks (TCNs) are able to address the limitations
of recurrent models by enabling parallel computation and stable gradient
propagation~\cite{pelletier2019} and are particularly suited for the
biological data analysis, where they can outperform attention-based models.
TCNs were first introduced as a generative audio model named WaveNet
~\cite{oord2016wavenet} and later appeared as an action
segmentation model in Lea et al.~\cite{lea2016temporal}.
    The general architecture proposed by Bai et al. \cite{bai2018empirical},
    however differs from such architectures by being much simpler.
    In fact, the model proposed by Bai, besides the basic concept of temporal convolutions listed before, uses only depth,
    dilation and residual connections to build the \textit{effective history} of the model.
    The effective history is defined as the ability of the networks to look into the past to make a prediction,
    or worded differently, it is the size of the context window.
    TCN have been successfully applied in several field of application, including classification of satellite image
    time series~\cite{pelletier2019}, clinical length of stay and mortality prediction~\cite{bednarski_temporal_2022}
    and energy-related time series forecasting~\cite{lara-benitez_temporal_2020}.

    Moreover, compared to attention-based approaches as Transformer based
    architectures \cite{vaswani2017attention},
    TCNs typically require fewer data and offer a
    favorable trade-off between model capacity and efficiency.
    Moreover, their convolutional backbone is naturally well-suited for
    modelling biological sequences.
    Key properties such as tunable receptive fields allow TCN-based
    architectures to capture long-range dependencies while maintaining architectural
    simplicity, making them especially amenable to downstream explainability analyzes.

%\subsection{Merging biological and computational perspectives}
%The mechanisms underlying TF binding and cooperation remain incompletely understood.

%Designing task-specific models that integrate state-of-the-art architectures with best
%practices in data efficiency is crucial for biological applications.
%Furthermore, the application of Explainable Artificial Intelligence (XAI)
%methods to trained models can help uncover biologically meaningful sequence features
%and shed light on the regulatory logic learned during training.
In the present study, %we aim for multiple TF DNA-binding site recognition as a
%multi-label classification problem, enabling the simultaneous prediction of binding events.
%More specifically, we aim to
we investigate whether deep learning models, as RNN and TCN,
can learn correlations
among TF labels solely from DNA sequence data, obtained from public repositories.
    In addition, we apply explainability methods to trained models to assess whether
    they can reveal biologically relevant sequence features and plausible interactions among TFs.
    Through this integrated biological and computational approach,
    we seek to gain new insights into the cooperative nature of transcriptional regulation.

    \section{Materials and Methods}

    We describe the datasets and the deep learning architectures we used
    in our analysis on Transcription Factor Binding Sites (TFBS).
    %\textcolor{red}{
        To start with, we describe separately the datasets we used for our multi-label TFBS recognition, that we created
        from raw ChIP-seq data for our study, and the one, not curated by us, that we used to benchmark our algorithms.
        Explicitly note that we used the formers to solve a multi-label classification problem while we used the latter
        to solve a binary classification problem, as it was created for that purpose (see \cite{zeng2016convolutional}).
    %}

    \subsection{Datasets for multi-label TFBS recognition}

   %\textcolor{red}{
       For the problem of multi-label TFBS recognition we constructed 3 datasets using publicly available ChIP-seq experiments
    provided by the ENCODE Consortium on the ENCODE portal~\cite{encode2012integrated}.
    In particular, we used all the ChIP-seq profiles available on the ENCODE portal matching the selection criteria we now describe.
% }
    Starting from MYC, a well-characterized TF with known cooperative behaviour,
    we followed two different approaches to determine which other TFs to include in the dataset.
The first approach yields the datasets we call \gzero and \gone.
We obtain them by selecting 5 and 7 additional TFs based on motif enrichment analysis
(SEA) in MYC-bound regions and data availability across 3 and 4 cell lines, respectively,
as depicted in Table~\ref{tab:g}, where the Group 0 represents \gzero, while the Group 1 represents \gone.
The second approach, yields the dataset we call H-M-E2F, obtained by
selected a hand-crafted set of TFs with putative interactions with MYC.
    More specifically we downloaded Chip-Seq targeting E2F1, E2F6, E2F8, MYC in K562 cell line from
    the ENCODE Regulation 'TF Clusters' track.
More specifically, the enrichment-driven approach resulted in the \textbf{\gzero} and \textbf{\gone} datasets
while the hand-curated selection of TFs resulted in the H-M-E2F dataset.

\begin{center}
    \begin{tikzpicture}[
        myrect/.style={
            rectangle,
            draw,
            inner sep=0pt,
            fit=#1}
    ]
        \node(table) {\begin{tabular}{l|rrrrrr}
 & K562 & GM12878 & HepG2 & H1-hESC & A549 & HeLa-S3 \\
TFs &  &  &  &  &  &  \\
\hline
ATF3 & 17244 & 1677 & 3291 & 4808 & 6580 & - \\
SP1 & 7206 & 18248 & 25477 & 15110 & - & - \\
TAF1 & 15246 & 14278 & 16659 & 20547 & 9984 & 16100 \\
USF2 & 3083 & 9022 & 6291 & 6952 & - & 12306 \\
c-Myc & 109625 & 3690 & 4413 & 5768 & - & 13061 \\
ELF1 & 27780 & 23008 & 18001 & - & 8611 & - \\
MAZ & 33323 & 18952 & 12090 & - & - & 13409 \\
IRF1 & 32550 & - & - & - & - & - \\
ETS1 & 10726 & 4120 & - & - & 5525 & - \\
ELK1 & 2961 & 5584 & - & - & - & 4809 \\
E2F4 & 8181 & 3440 & - & - & - & 2831 \\
IRF4 & - & 17771 & - & - & - & - \\
SP2 & 3124 & - & 2626 & 2469 & - & - \\
CTCF & - & - & 191734 & 171742 & 180057 & 149989 \\
IRF3 & - & - & 684 & - & - & 1587 \\
ELK4 & - & - & - & - & - & 5916 \\
STAT2 & 4963 & - & - & - & - & - \\
ATF2 & - & 23467 & - & 5998 & - & - \\
SP4 & - & - & - & 5752 & - & - \\
\end{tabular}
};

% Rettangolo blu (Group 1)
        \node(g1) [
        draw, blue, thick, rounded corners,
        label=left:\textcolor{blue}{Group 1},
        fit={
            ($(table.north west)!.11!(table.south west)$)
            ($(table.north)!.11!(table.south)!.45!(table.east)$)
        }
        ]{};

% Rettangolo rosso (Group 0)
        \node(g0) [
        draw, red, thick, rounded corners,
        label=left:\textcolor{red}{Group 0},
        fit={
            ($(table.north west)!.12!(table.south west)$)
            ($(table.center)!.23!(table.north east)!.27!(table.east)$)
        }
        ]{};
%  ($(table.south) !.3! (table.north)$)
%  rectangle
%  ($(table.south east) !.7! (table.north east)$);
% \path   ($(table.south) !.3! (table.north)$) -- ($(table.south east) !.7! (table.north east)$) node[midway,above] {$m$};

    \end{tikzpicture}
    \captionof{table}{ChIP-seq peaks available for each TF with respect to the cell line. \textcolor{red}{ Group 0} and \textcolor{blue}{ Group 1} are noted in color.}
    \label{tab:g}
\end{center}

%{\color{blue} Samples of data in \gzero, \gone and H-M-E2F are found in WEBSITE.}

 %    \textcolor{red}{ Questa parte ora mi sembra superflua e la rimuoverei \\
 %   For all ChIP-seq, peaks were clustered based on genomic overlap, and sequences of
 %   1000 bp centered on cluster midpoints were extracted from the hg19 reference genome.
 %   Each sequence was labeled with all TFs whose peaks contributed to the cluster.
 %   }

%\textcolor{blue}{Due to the inherent imbalance in TF binding events, we opted against resampling techniques to preserve biological signal. Instead, we applied class-weighted binary cross-entropy loss during training. }
    % We collected available ChIP-seq profiles from ENCODE \cite{encode2012integrated} portal for a set of Transcription Factors (TF) of interest.
    % Next, we performed peak intersection between the collected profiles. Finally, in order to generate our final dataset we collect slices of $1000bp$ centered on the mid-point of each resulting peak overlap. The resulting sequences are labeled as being bound by the TFs associated to the corresponding peaks.

    We collected ChIP-seq profiles for selected TFs from the ENCODE portal~\cite{encode2012integrated}.
    Peak intersections were then computed across selected profiles.
    For each overlapping region, we extracted a $1000 bp$ sequence centred at the midpoint.
    Each sequence was encoded with one-hot encoding and labelled with the TFs corresponding
    to the intersecting peaks.

    The datasets thus obtained are then a set of sequence features $X$ and a set of labels $Y$
    where:
    \[
                                x_i \in X, 0 \leq i \leq |X|
    \]
    represents the sequence vector of the $i^{th}$ sample and
    \[
        y_i = [l^0_i , ..., l^k_i ] \in Y, 0 \leq i \leq |Y|
    \]
    represents the label vector consisting of $k$ binary variables
    $l^k_i$ encoding the presence/absence of the $k$ TFs, as usual $|X|$ and $|Y|$ denoting
    the numerosity of samples in the dataset $X$ with labels $Y$.
    It is worth noting that while in the binary classification setting we distinguish
    between bound (positive) and unbound (negative), in our main formalization there is no
    absolute negative or unbound label; in fact each sequence example is labelled with one or
    more TFs.
    We train our proposed deep-learning models to jointly learn the probability of
    the label vector $y_i$ modelling each $l^k_i$  as individual binary predictions,
    therefore predicting  binding specificities for all TFs simultaneously.

    % \begin{multicols}{2}
    %    \begin{column}{\widthfrac{1}{3}}
    % \begin{figure}[h!]
   %     \centering
   %     \resizebox{0.5\textwidth}{!}
        {
   %     \input{aux_files/graphs/TCN_model}
        %\medskip
        %\input{aux_files/graphs/TCN_model}
        }
        % \captionof{figure}{Baseline sequence encoder composed by two 1-D CNN layers and one Bi-LSTM layer.}
 %       \label{fig:TCN-model}

 %   \end{figure}
    %     \end{column}
    %     \begin{column}{\widthfrac{2}{3}}
{%\color{red}
\subsection{Dataset for binary TFBS recognition}\label{sec:binlab}
To benchmark our algorithms we used the dataset curated by Zeng et al. ~\cite{zeng2016convolutional}, which includes 165 ChIP-seq
datasets selected by the authors of \textit{op. cit.} ensuring diversity across cell lines.
The source of this data is the ENCODE consortium, the same source of data we used to create our multi-label dataset.
This curated dataset was provided to us by the authors of  ~\cite{zeng2016convolutional} and was already labeled by them.
}
    Each dataset was provided to us already split into train ($80\%$) and test ($20\%$) sets
    with positive and negative instances.
    During training and development we further spilt each train set to obtain a validation set consisting of $20\%$ the
    initial train set.
    DNA sequences are 101 bp long and labeled binarily to indicate Transcription
    Factor Binding Sites (TFBS) presence.

    \subsection{Deep Learning Architectures}
\begin{figure}[ht!]
        \centering
        \begin{subfigure}{.5\textwidth}
            \centering
            \resizebox{0.9\textwidth}{!}
            {
            %\begin{figure}[h!]
\centering
\begin{tikzpicture}
\tikzstyle{RNNunit}=[draw,shape=rectangle,minimum size=0.5cm, fill opacity = 1, fill= black!5!white]
\tikzstyle{ldots} = [draw,shape=rectangle,minimum size=0.5cm, draw opacity = 0]
\tikzstyle{unit}=[draw,shape=circle,minimum size=0.05cm]
\centering

    \pgfmathsetmacro{\rnnstart}{6};
    \pgfmathsetmacro{\rnnmid}{5};
    \pgfmathsetmacro{\rnnend}{0};

    \pgfmathsetmacro{\cnnonestart}{9.5};
    \pgfmathsetmacro{\cnnoneend}{9};
    \pgfmathsetmacro{\cnntwostart}{8.5};
    \pgfmathsetmacro{\cnntwoend}{8};
    \pgfmathsetmacro{\cnnthreestart}{7.5};
    \pgfmathsetmacro{\cnnthreeend}{7};
    \pgfmathsetmacro{\cnnwidth}{2};

    \pgfmathsetmacro{\tcnonestart}{6.5};
    \pgfmathsetmacro{\tcnoneend}{6};
    \pgfmathsetmacro{\tcntwostart}{5.5};
    \pgfmathsetmacro{\tcntwoend}{5};
    \pgfmathsetmacro{\tcnthreestart}{4.5};
    \pgfmathsetmacro{\tcnthreeend}{4};
    \pgfmathsetmacro{\tcnwidth}{2};

    \pgfmathsetmacro{\mlpstart}{2};
    \pgfmathsetmacro{\mlpmid}{-2.5};
    \pgfmathsetmacro{\mlpend}{0.5};

    \node (dnk) at (0,13.5) {\tt GTGCATCTGACTCCTGAGGAGTAG};
    % \node (onehot) at (0,15) {\includegraphics[width=1.2\textwidth]{mygraphs/onehot.png}};

	% \node at (-2.7, 14) {\tt \dots};
	% \node at (2.7, 14) {\tt \dots};
	% \draw[red, opacity=0.4, very thick] (-3, 0) -- (3, 0);

	\node at (5, 13.5) {DNA};
		
    \node (rnk) at (0,12){};
    % \draw[fill=black,fill opacity=0.1,draw=black] (3,11.5) rectangle (-3,11);

    \path[fill stretch image=aux_files/graphs/onehot, draw=black] (3,12) rectangle (-3,11);
    
    \draw[-stealth, thick] (dnk) -- node[right] {} (rnk);
    \node at (5, 11.25) {Embedding layer};
    
    \coordinate (SE-top-left) at (-3,10);
    \coordinate (SE-bottom-right) at (3, \rnnmid -2);
    \draw[fill= black!60!green,fill opacity=0.1,draw=black!60!green, on background layer] (SE-top-left) rectangle (SE-bottom-right);
    \draw[-stealth, thin] (0, 11) -- node[right] {} (0, 10);
    % \draw[-stealth, thin] (0, \rnnmid -2) -- node[right] {} (0, \rnnmid -3);
    
    % CNN

    % Layers

    \draw[fill= black!5!white,opacity=0.1,draw=black, opacity = 1] (\cnnwidth,\cnntwostart) rectangle   (-\cnnwidth,\cnntwoend);
    \draw[fill= black!5!white,opacity=0.1,draw=black, opacity = 1] (\cnnwidth,\cnnonestart) rectangle   (-\cnnwidth,\cnnoneend);
    \draw[fill= black!5!white,opacity=0.1,draw=black, opacity = 1] (\cnnwidth,\cnnthreestart) rectangle   (-\cnnwidth,\cnnthreeend);

    \node at ($(5, \cnnonestart) !.5! (5, \cnnthreeend) $) {\small CNN layers};

     % \node at ($(5.5, \mlpmid) !.5! (5.5, \mlpend) $) {Layer 2};
   %  \draw[fill= black!5!white,opacity=0.1,draw=black, opacity = 1] (\cnnwidth,\cnntwostart) rectangle   (-\cnnwidth,\cnntwoend);
   % \draw[fill=black,opacity=0.1,draw=black, draw opacity = 1] (\cnnwidth,\cnnonestart) rectangle   (-\cnnwidth,\cnnoneend);
    % \draw[fill=black,opacity=0.1,draw=black, draw opacity = 1] (\cnnwidth,\cnnthreestart) rectangle   (-\cnnwidth,\cnnthreeend);

    % Conv Operations
    
    \draw[fill=red,opacity=0.2,draw=red, draw opacity = 1](-\cnnwidth,\cnnonestart) rectangle   (-\cnnwidth +1.5,\cnnoneend);
    \draw[fill=red,opacity=0.2,draw=red, draw opacity = 1](-\cnnwidth,\cnntwostart) rectangle   (-\cnnwidth +0.5,\cnntwoend);

    \draw[draw = red,opacity=0.2]
    (-\cnnwidth,\cnnoneend) -- (-\cnnwidth ,\cnntwostart)
    (-\cnnwidth +1.5,\cnnoneend) -- (-\cnnwidth +0.5 ,\cnntwostart);

    \draw[fill=red,opacity=0.2,draw=red, draw opacity = 1](-\cnnwidth +2,\cnntwostart) rectangle   (-\cnnwidth +1.5+2,\cnntwoend);
    \draw[fill=red,opacity=0.2,draw=red, draw opacity = 1](-\cnnwidth +2,\cnnthreestart) rectangle   (-\cnnwidth +2+0.5,\cnnthreeend);

    \draw[draw = red,opacity=0.2]
    (-\cnnwidth +2,\cnntwoend) -- (-\cnnwidth +2,\cnnthreestart)
     (-\cnnwidth +1.5+2,\cnntwoend) --  (-\cnnwidth +2+0.5,\cnnthreestart);

    % TCN 

    % Layers
    
    \draw[fill= black!5!white,opacity=0.1,draw=black, opacity = 1] (\tcnwidth,\tcntwostart) rectangle   (-\tcnwidth,\tcntwoend);
    \draw[fill= black!5!white,opacity=0.1,draw=black, opacity = 1] (\tcnwidth,\tcnonestart) rectangle   (-\tcnwidth,\tcnoneend);
    \draw[fill= black!5!white,opacity=0.1,draw=black, opacity = 1] (\tcnwidth,\tcnthreestart) rectangle   (-\tcnwidth,\tcnthreeend);

    \node at ($(5, \tcnonestart) !.5! (5, \tcnthreeend) $) {\small TCN layers};

    % Conv Operations
    
    \draw[fill=blue,opacity=0.2,draw=blue, draw opacity = 1](-\tcnwidth,\cnnthreestart) rectangle   (-\tcnwidth +1.5,\cnnthreeend);
    \draw[fill=blue,opacity=0.2,draw=blue, draw opacity = 1](-\tcnwidth +1,\tcnonestart) rectangle   (-\tcnwidth +1.5,\tcnoneend);
    
    \draw[draw = blue,opacity=0.2]
    (-\tcnwidth,\cnnthreeend) -- (-\tcnwidth +1,\tcnonestart)
    (-\tcnwidth +1.5,\cnnthreeend) -- (-\tcnwidth +1.5 ,\tcnonestart);

    \draw[fill=blue,opacity=0.2,draw=blue, draw opacity = 1](0,\tcnonestart) rectangle   (0 +1.5,\tcnoneend);
    \draw[fill=blue,opacity=0.2,draw=blue, draw opacity = 1](0 +1,\tcntwostart) rectangle   (0 +1.5,\tcntwoend);
    
    \draw[draw = blue,opacity=0.2]
    (0,\tcnoneend) -- (0 +1,\tcntwostart)
    (0 +1.5,\tcnoneend) -- (0 +1.5 ,\tcntwostart);

    \draw[fill=blue,opacity=0.2,draw=blue, draw opacity = 1](-\tcnwidth +1,\tcntwostart) rectangle   (-\tcnwidth +1.5 +1,\tcntwoend);
    \draw[fill=blue,opacity=0.2,draw=blue, draw opacity = 1](-\tcnwidth +2,\tcnthreestart) rectangle   (-\tcnwidth +1.5 +1,\tcnthreeend);
    
    \draw[draw = blue,opacity=0.2]
    (-\tcnwidth +1,\tcntwoend) -- (-\tcnwidth +1 +1 ,\tcnthreestart)
    (-\tcnwidth +1.5 +1,\tcntwoend) -- (-\tcnwidth +1.5 +1 ,\tcnthreestart);

   % \draw[-stealth, thin, opacity =0.5] ($(x3.north east) !.2! (x3.south east) $) -- node[right] {} ($ (x2.north west) !.2! (x2.south west) $);
 %   \draw[-stealth, thin, opacity =0.5] ($(x3.north east) !.8! (x3.south east) $) -- node[right] {} ($ (x2.north west) !.8! (x2.south west) $);
    % \draw[-stealth, thin, opacity =0.5] (x1.north east) -- node[right] {} (x0.north west);
    % \draw[-stealth, thin, opacity =0.5] (x1. east) -- node[right] {} (x0. west);
    \draw[thin] (\tcnwidth -0.5 ,\tcnthreestart) --  (\tcnwidth -0.5 , \tcnthreeend) ;

    \draw[-|, thin] (\tcnwidth -0.25 ,\tcnthreeend)  -- (\tcnwidth -0.25 ,\tcnthreeend -0.25) -- (0 ,\tcnthreeend -0.25) -- (0 ,\tcnthreeend -0.5);

    % MLP 

    \node at (0,\mlpstart){\ldots};

    \foreach \i in {0,1,2,3,5,6,7,8}
{
        \pgfmathsetmacro{\y}{ \mlpstart }; 
        \pgfmathsetmacro{\x}{2.75 - (\i *  0.6875)};
        % \pgfmathtruncatemacro{\label}{x\i};
        \node[unit](xxx\i) at (\x, \y) {};
        \draw[-stealth, thin] (\x, \mlpstart + 1) -- node[right] {} (xxx\i);
        
}

    \foreach \i/\l in {0/E2F1,1/E2F6,2/E2F8,3/MYC}
    {
        \pgfmathsetmacro{\y}{ \mlpend }; 
        \pgfmathsetmacro{\x}{2.325 - (\i * 1.55)};
        % \pgfmathtruncatemacro{\label}{xxx\i};
        \node[unit][label=below:\l](xxxx\i) at (\x, \y) {};
        %\draw[-stealth] (x\i) -- node[right] {} (xx\i);
        
}

    \foreach \i in {0,1,2,3,5,6,7,8}{
           \foreach \j in {0,...,2,3}{
                \draw[-stealth, thin, line width=0.1pt,opacity=0.5] (xxx\i) -- node[right] {} (xxxx\j);
            
            }
    }

    \node at ($(5, \mlpstart) !.5! (5, \mlpend) $) {Classifier};
    
\end{tikzpicture}
            }
            \caption{TCN-based model}
            \label{fig:sub1}
        \end{subfigure}%
        \begin{subfigure}{.5\textwidth}
            \centering
            \resizebox{0.9\textwidth}{!}
            {
            %\begin{figure}[h!]
\centering
\begin{tikzpicture}
\tikzstyle{RNNunit}=[draw,shape=rectangle,minimum size=0.5cm, fill opacity = 1, fill= black!5!white]
\tikzstyle{ldots} = [draw,shape=rectangle,minimum size=0.5cm, draw opacity = 0]
\tikzstyle{unit}=[draw,shape=circle,minimum size=0.05cm]
\centering

    \pgfmathsetmacro{\rnnstart}{6};
    \pgfmathsetmacro{\rnnmid}{5};
    \pgfmathsetmacro{\rnnend}{0};

    \pgfmathsetmacro{\cnnonestart}{9.5};
    \pgfmathsetmacro{\cnnoneend}{9};
    \pgfmathsetmacro{\cnntwostart}{8.5};
    \pgfmathsetmacro{\cnntwoend}{8};
    \pgfmathsetmacro{\cnnthreestart}{7.5};
    \pgfmathsetmacro{\cnnthreeend}{7};
    \pgfmathsetmacro{\cnnwidth}{2};

    \pgfmathsetmacro{\mlpstart}{2};
    \pgfmathsetmacro{\mlpmid}{-2.5};
    \pgfmathsetmacro{\mlpend}{0.5
    };

    \node (dnk) at (0,13.5) {\tt GTGCATCTGACTCCTGAGGAGTAG};
    % \node (onehot) at (0,15) {\includegraphics[width=1.2\textwidth]{mygraphs/onehot.png}};

    % \node at (-2.7, 14) {\tt \dots};
    % \node at (2.7, 14) {\tt \dots};
    % \draw[red, opacity=0.4, very thick] (-3, 0) -- (3, 0);

    \node at (5, 13.5) {DNA};

    \node (rnk) at (0,12){};
    % \draw[fill=black,fill opacity=0.1,draw=black] (3,11.5) rectangle (-3,11);

    \path[fill stretch image=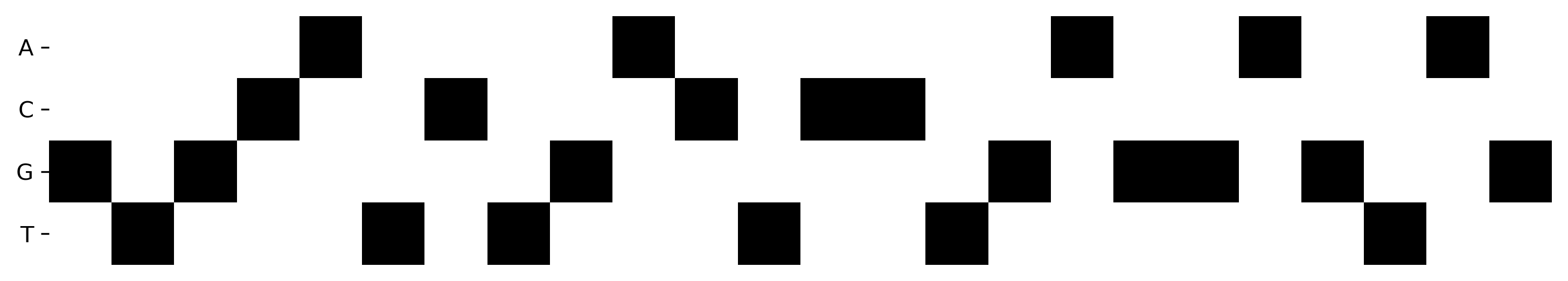, draw=black] (3,12) rectangle (-3,11);

    \draw[-stealth, thick] (dnk) -- node[right] {} (rnk);
    \node at (5, 11.25) {Embedding layer};

    \coordinate (SE-top-left) at (-3,10);
    \coordinate (SE-bottom-right) at (3, \rnnmid -2);
    \draw[fill= black!60!green,fill opacity=0.1,draw=black!60!green, on background layer] (SE-top-left) rectangle (SE-bottom-right);
    \draw[-stealth, thin] (0, 11) -- node[right] {} (0, 10);
    % \draw[-stealth, thin] (0, \rnnmid -1.5) -- node[right] {} (0, \rnnmid -2.5);
    % CNN

    % Layers

    \draw[fill= black!5!white,opacity=0.1,draw=black, opacity = 1] (\cnnwidth,\cnntwostart) rectangle   (-\cnnwidth,\cnntwoend);
    \draw[fill= black!5!white,opacity=0.1,draw=black, opacity = 1] (\cnnwidth,\cnnonestart) rectangle   (-\cnnwidth,\cnnoneend);
    \draw[fill= black!5!white,opacity=0.1,draw=black, opacity = 1] (\cnnwidth,\cnnthreestart) rectangle   (-\cnnwidth,\cnnthreeend);

    \node at ($(5, \cnnonestart) !.5! (5, \cnnthreeend) $) {\small CNN layers};

    % Conv Operations
    
    \draw[fill=red,opacity=0.2,draw=red, draw opacity = 1](-\cnnwidth,\cnnonestart) rectangle   (-\cnnwidth +1.5,\cnnoneend);
    \draw[fill=red,opacity=0.2,draw=red, draw opacity = 1](-\cnnwidth,\cnntwostart) rectangle   (-\cnnwidth +0.5,\cnntwoend);

    \draw[draw = red,opacity=0.2]
    (-\cnnwidth,\cnnoneend) -- (-\cnnwidth ,\cnntwostart)
    (-\cnnwidth +1.5,\cnnoneend) -- (-\cnnwidth +0.5 ,\cnntwostart);

    \draw[fill=red,opacity=0.2,draw=red, draw opacity = 1](-\cnnwidth +2,\cnntwostart) rectangle   (-\cnnwidth +1.5+2,\cnntwoend);
    \draw[fill=red,opacity=0.2,draw=red, draw opacity = 1](-\cnnwidth +2,\cnnthreestart) rectangle   (-\cnnwidth +2+0.5,\cnnthreeend);

    \draw[draw = red,opacity=0.2]
    (-\cnnwidth +2,\cnntwoend) -- (-\cnnwidth +2,\cnnthreestart)
     (-\cnnwidth +1.5+2,\cnntwoend) --  (-\cnnwidth +2+0.5,\cnnthreestart);

    % RNN

    \node[ldots] (x2) at (0,\rnnstart){\ldots} ;
        \foreach \i in {0,1,3,4}{

        \pgfmathtruncatemacro{\y}{ \rnnstart }; 
        \pgfmathsetmacro{\x}{1.75 - (\i *  0.875)};

        \node[RNNunit](x\i) at (\x, \y) {};
        \draw[-stealth, thin] (\x, \rnnstart + 1) -- node[right] {} (x\i);
        
    }

     \foreach \i in {4,3,2,1}
{
       \pgfmathtruncatemacro{\j}{ \i -1};
    \draw[-stealth, thin, opacity =0.5] ($(x\i.north east) !.2! (x\i.south east) $) -- node[right] {} ($ (x\j.north west) !.2! (x\j.south west) $);
    \draw[-stealth, thin, opacity =0.5] ($(x\i.north east) !.8! (x\i.south east) $) -- node[right] {} ($ (x\j.north west) !.8! (x\j.south west) $);
}

    \node[ldots] (xx2) at (0,\rnnmid){\ldots};
     \foreach \i in {0,1,3,4}{
        \pgfmathtruncatemacro{\y}{ \rnnmid }; 
        \pgfmathsetmacro{\xx}{1.75 - (\i *  0.875)};
        % \pgfmathtruncatemacro{\label}{xx\i};
        \node[RNNunit](xx\i) at (\xx, \y) {};
         \draw[-stealth, thin] (\xx,  \rnnstart +0.5) --  (\xx -0.45,  \rnnstart + 0.5) -- (\xx -0.45, \rnnmid +0.5) --   (\xx, \rnnmid +0.5) -- (xx\i);
    }

     \foreach \i in {0,1,2,3}{
       \pgfmathtruncatemacro{\j}{ \i +1};
    \draw[-stealth, thin, opacity =0.5] ($(xx\i.north west) !.2! (xx\i.south west) $) -- node[right] {} ($ (xx\j.north east) !.2! (xx\j.south east) $);
    \draw[-stealth, thin, opacity =0.5] ($(xx\i.north west) !.8! (xx\i.south west) $) -- node[right] {} ($ (xx\j.north east) !.8! (xx\j.south east) $);
}

     \draw[-|, thin] (xx4.south) -- (-1.75, \rnnmid-0.5) --(0, \rnnmid-0.5) (x0.south)--(1.75 , \rnnstart-0.4) -- (1.75 +0.45, \rnnstart-0.4) -- (1.75 +0.45, \rnnmid-0.5) --(0, \rnnmid-0.5) --(0, \rnnmid-1) ;

    \node at ($(5, \rnnstart) !.5! (5, \rnnmid) $) {\small Bi-LSTM layer};

    % MLP 

    \node at (0,\mlpstart){\ldots};

    \foreach \i in {0,1,2,3,5,6,7,8}
        {
        \pgfmathsetmacro{\y}{ \mlpstart };
        \pgfmathsetmacro{\x}{2.75 - (\i *  0.6875)};
        % \pgfmathtruncatemacro{\label}{x\i};
        \node[unit](xxx\i) at (\x, \y) {};
        \draw[-stealth, thin] (\x, \mlpstart + 1) -- node[right] {} (xxx\i);

    }

    \foreach \i/\l in {0/E2F1,1/E2F6,2/E2F8,3/MYC}
        {
        \pgfmathsetmacro{\y}{ \mlpend };
        \pgfmathsetmacro{\x}{2.325 - (\i * 1.55)};
        % \pgfmathtruncatemacro{\label}{xxx\i};
        \node[unit][label=below:\l](xxxx\i) at (\x, \y) {};
        %\draw[-stealth] (x\i) -- node[right] {} (xx\i);

    }

    \foreach \i in {0,1,2,3,5,6,7,8}{
        \foreach \j in {0,...,2,3}{
            \draw[-stealth, thin, line width=0.1pt,opacity=0.5] (xxx\i) -- node[right] {} (xxxx\j);

        }
    }

    \node at ($(5, \mlpstart) !.5! (5, \mlpend) $) {Classifier};

\end{tikzpicture}}
            \caption{Hybrid CNN-RNN baseline}
            \label{fig:sub2}
        \end{subfigure}%
        \caption{Diagrams of the implemented models' architecture.}
        \label{fig:models}
    \end{figure}
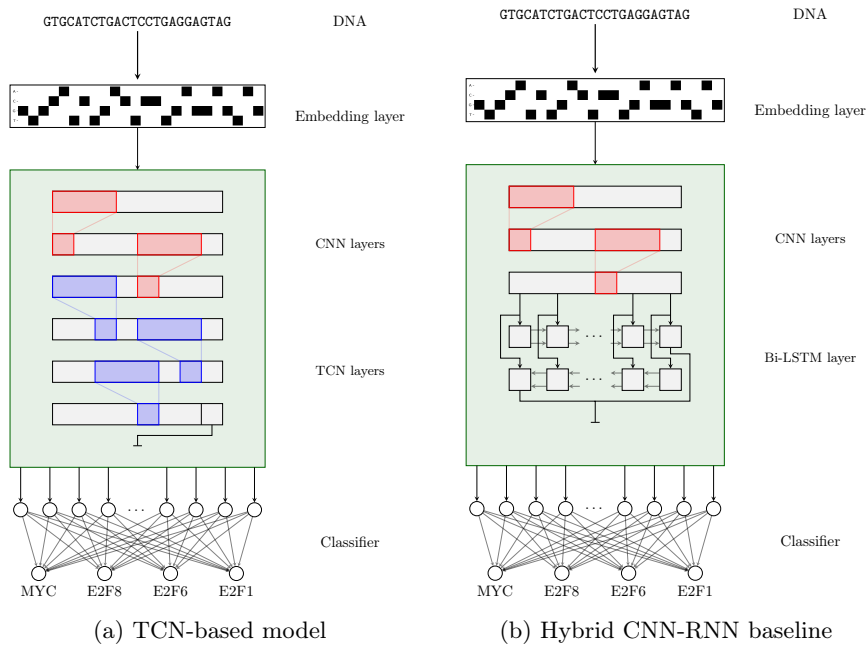

    %We developed and tested several Deep Learning (DL) architectures to address the \textcolor{Decepticon purple}{TF binding prediction task formalized as a multi-label}\footnote{Multi-label classification involves predicting zero or more non-exclusive class labels for each example.} problem.
    We designed and evaluated several Deep Learning (DL) models to address the
    main goal of this work, namely TF binding prediction task as a multi-label problem.
    %More specifically we developed an Recurrent Neural Network (RNN) based baseline as well as a Temporal Convolutional Network (TCN) based model.
    We implemented a Temporal Convolutional Network(TCN)-based model alongside a hybrid baseline model based upon Recurrent Neural Network(RNN) and Convolutional Neural Networks(CNN).
    The baseline model is almost identical to the TCN model except for the TCN blocks which are substituted by two layers of Bi-LSTM with \textit{hidden\_dimention = 50}.
    A visual representation can be found in Figure~\ref{fig:models} as well as additional architecture details in Table~\ref{tab:Hyper Parameters} .
% \begin{block}{Why TCNs?
%We now describe more in detail the different DL architectures.
%\par{\bf Temporal Convolutional Networks.}
    TCNs are convolutional architectures for sequence modelling~\cite{bai2018empirical},
    which have demonstrated strong performance across different domains.
    \noindent  TCNs' convolutional backbone enables effective modelling of long-range dependencies, arguably granting them an advantage in comparison with RNNs.
    In addition, convolutional architectures are characterized by a strong inductive bias suitable for local pattern detection, making them particularly effective for biological sequence analysis. \\
    \noindent TCNs have two distinguishing features:
    \begin{itemize}
        \item All the convolutions operations in the network are causal, there is no information flowing from future to past.
        \item The architecture can take as input a sequence of arbitrary length and map it to an output sequence of the same length similarly to RNN\@.
    \end{itemize}

    \noindent We will now introduce briefly  the main features of the TCN architecture proposed by Bai~\cite{bai2018empirical} which we used as the foundations to develop our temporal convolutional models.

    \par
\begin{figure}[!h]
        \centering
        \includegraphics[width=0.5\linewidth]{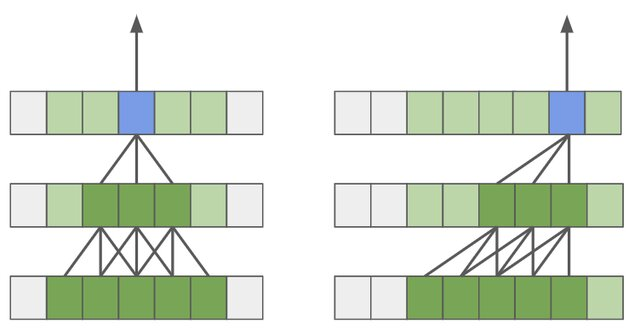}
        \caption{Causal convolutions by the use of padding; on the left 1D convolution with "valid" padding, on the right 1D convolution with left padding enforcing causality
        \cite{kondratyuk2021movinets}.}
        \label{fig:casual_comb}
    \end{figure}

{\textbf{Causal Convolutions.}}
    The classic convolution layer is rendered causal by appropriate use of padding.
    In fact, the convolution operation is allowed access only to past events by zero-padding asymmetrically the sequence at the start.
    The padding , thus, ensures that there is no information leakage from future elements of the sequences as show in Fig.~\ref{fig:casual_comb}.

    \noindent The choice of appropriate padding also ensures that the output size matches the input size of the sequence.

    \par{\textbf{Dilated Convolutions.}}
    The effective history grows linearly with the depth of the network,  making it challenging to achieve a good enough context window for longer sequences.
    The effective history of the network is dependant also on the convolutions' receptive field;
    due to this reason dilated convolutions represent an ideal solution to this challenge.
    Dilated convolutions in fact, enable an exponentially larger receptive field compared to classic convolutions~\cite{yu2015multi},
    thus increasing the effective history of the network without increasing its depth.
    The use of dilated convolution in TCN has been first introduced in~\cite{oord2016wavenet}.

    \noindent More formally a dilated convolution is defined as:

    \[
        (x*_dF)(s) = \sum_{i = 0}^{k-1}F(i)\cdot x_{s-d\cdot i}
    \]

    \noindent Where $d$ represents the dilation factor and $k$ represents the kernel size.
    When $d=1$ dilation convolution and regular convolution are equivalent.

    \par{\textbf{Residual Connections.}}
    Residual blocks where first introduced by He et al. \cite{he2016deep}.
    Residual connections works by adding an identity mapping parallel to the convolutional block and then summing it with the output.
    More formally, for a generic layer $F$ with input $x$ the output $O$ of the residual block can be defined as:

    \[
        O = \sigma(F(x)+x)
    \]
    Where $\sigma(.)$ is the activation function of choice.
    It is worth noting that $x$ and $F(x)$ might not have the same
    dimensions, in which case a linear projection is added.

    \begin{figure}[h!]
        \centering
        \includegraphics[width=0.5\linewidth]{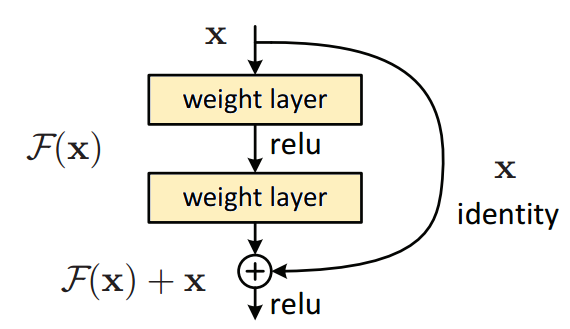}
        \caption{A residual block}
        \label{fig:res_block}
    \end{figure}

    \noindent Residual blocks allow the layers to learn just the residual feature map while the previous one is carried by the identity connection.
    Empirical results show that convolutional layers perform much better by learning on the residual feature map, rather than on the whole feature map itself.
    This enables layers to learn modifications to the identity mapping instead of the full transformation.
    In addition, residual connections stabilize larger networks by allowing an easier propagation for a fine-tuned signal;
    in fact, the unmodified propagation requires just setting all the layer's weights to 0, which is much simpler than learning the identity operation.
    Due to the dependence of the temporal convolution's effective history on the depth of the network, residual connections represent an optimal architecture for TCNs, as it allows for  deeper architectures.
    \begin{figure}[!h]
        \centering
        \includegraphics[width=\linewidth]{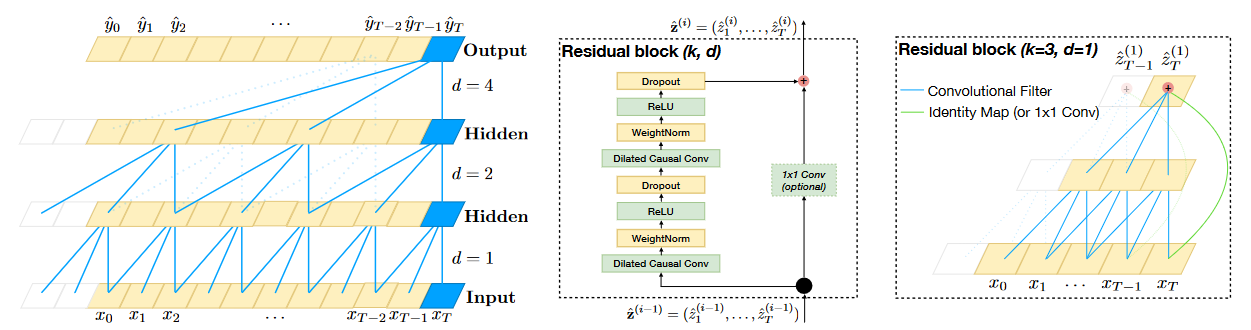}
        \caption{Temporal Convolutional Networks as proposed by Bai et al.\cite{bai2018empirical}.}
        \label{fig:tcn}
    \end{figure}

\begin{table}[h!]
    \centering
    \begin{tabular}{l|l|l|l}
        \textbf{Dataset} & HC-M-E2F &
        D-7TF-2CL & D-5TF-3CL
        \\
        \hline
        \textbf{Parameter} & \textbf{Final Value} & \textbf{Final Value} & \textbf{Final Value} \\
        \hline
        batch size & 64 & 64 & 64 \\
        dropout ratio & 0.5 & 0.5 & 0.5 \\
        % early stopping & True & True & True \\
        epochs & 50 & 50 & 50 \\
        learning rate & 0.00258 &
        0.00508 & 0.00219 \\
        % optimizer & Adam & Adam & Adam \\
        MLP hidden size & 100 & 100 & 100 \\
        CNN kernel number & 32 & 32 & 32 \\
        CNN layers & 2 & 2 & 2 \\
        TCN kernel size & 32 & 32 & 32 \\
        TCN block number & 6 & 5 & 6 \\
    \end{tabular}
    \caption{Model Hyperparameters}
    \label{tab:Hyper Parameters}
\end{table}

    \subsection{Attribution Methods}\label{subsec:attribution-methods-and-implementation}

    %In order to make gain insight on the features learned by the trained models (DNA patterns) we employed several explainability techniques.
    %We first obtained the attributions scores for each sequence which quantify the importance of each nucleotide in the sequence on the final model's output.
    %We obtained the attribution scores for the trained model with rthe use of Integrated Gradients \cite{sundararajan2017axiomatic}.
    %Integrated Gradients is an attribution technique which considers also a baseline distribution for computing feature attribution \cite{reviewdevastantecontuteleattrib}. In the present project we adopted as baselines shuffled sequences obtained via di-nucleotide shuffling.
    %We computed attributions scores for each of the target TFs for each sequence separately.
    %Next we employed TF-MODISCO \cite{MODISCO} to extract relevant seq-lets from the computed attributions.

    To gain insight into the DNA patterns learned by the models, we applied several explainability techniques.
    % \textcolor{Decepticon purple}{
    Attribution scores that quantify the contribution of each nucleotide %}
    to the model’s output were computed using Integrated Gradients~\cite{sundararajan2017axiomatic}, with di-nucleotide shuffled sequences as baselines.
Attribution was performed separately for each target TF. Subsequently, we used TF-MoDISco~\cite{shrikumar2018technical}
to identify and extract informative seqlets, short genomic sequences with high information content, from the attribution maps.

\subsection{Implementation}
We implemented our models with the \texttt{pytorch}\cite{ansel2024pytorch} framework performing hyperparameter tuning using Tree of Parzen Estimators~\cite{bergstra2011algorithms} provided by the \texttt{hyperopt} library~\cite{bergstra2013making}.
In addition, we also employed MLFlow~\cite{zaharia2018accelerating}, Pandas~\cite{reback2020pandas}, scikit-learn~\cite{scikit-learn} and numpy~\cite{harris2020array} python libraries during the development of our models and training scripts.
We trained all our models with the Adam optimizer with weight decay set to 0 in conjunction with a custom learning rate scheduler.
The scheduler is composed of a linear warmup phase for the first 20\% of training epochs followed by cosine annealing to mitigate overfitting.
During training we also employed early stopping with a patient mechanism.
    The final hyperparameters can be found in Table~\ref{tab:Hyper Parameters}.

      %  \input{aux_files/tabs/model_info}
%    \newpage % da rimuover per impaginare bene
    \section{Results and discussion}

    We detail our results on TF binary classification, as sanity check and benchmarking
    and then as multi-label TF binding sites classification, the main goal of our work.

\subsection{Binary classification as benchmarking}
In order to assess the general capabilities of the TCN architecture on the classification of
genetic sequences we also tested our model in the binary classification setting.
% \textcolor{red}{
    To this end, we used the dataset described in Section \ref{sec:binlab} that has been widely used for DNA–TF binding site prediction
\cite{zhang2024mlsnet, ding2023deepstf}.
%}
%Each dataset was provided to us already split into train ($80\%$) and test ($20\%$) sets
%with positive and negative instances.
%During training and development we further spilt each train set to obtain a validation set consisting of $20\%$ of the
%initial train set.
%DNA sequences are 101 bp long and labeled binarily to indicate Transcription
%Factor Binding Sites (TFBS) presence.

  %  \begin{figure}
  %      \centering
  %      \includegraphics[width=0.75\linewidth]{aux_files/graphs/violin}
  %      \caption{\textcolor{red}{Questa la rimoverei}}
  %      \label{fig:violinplot}
  %  \end{figure}

    \begin{table}[ht]
\centering
\begin{tabular}{l|rrr}
\hline
 & AP & AU-ROC & Accuracy\\
\hline
% count & 165.00 & 165.00 \\
mean & 0.88 & 0.87 & 0.80 \\
std & 0.11 & 0.11 & 0.11\\
min & 0.49 & 0.49 & 0.48\\
25\% & 0.84 & 0.83 & 0.75\\
50\% & 0.91 & 0.90 &  0.82\\
75\% & 0.96 & 0.95 & 0.89 \\
max & 0.99 & 0.99 & 0.95\\
\hline
\end{tabular}
\caption{Descriptive statistiscs for the distributions of AP, AU-ROC and Accuracy across the 165 binary datasets.}\label{tab:binsum}
\end{table}

%\begin{figure}
%        \centering
%\includegraphics[width=1\linewidth]{aux_files/graphs/binary_metrics_violin}
%        \caption{Violin plot of AUC, AP and accuracy across the 165 binary datasets \textcolor{red}{Terrei solo o questa o la \ref{fig:bin_alltogheter}}}
%        \label{fig:bin_violin}
%    \end{figure}

    \begin{figure}
        \centering
        \includegraphics[width=1\linewidth]{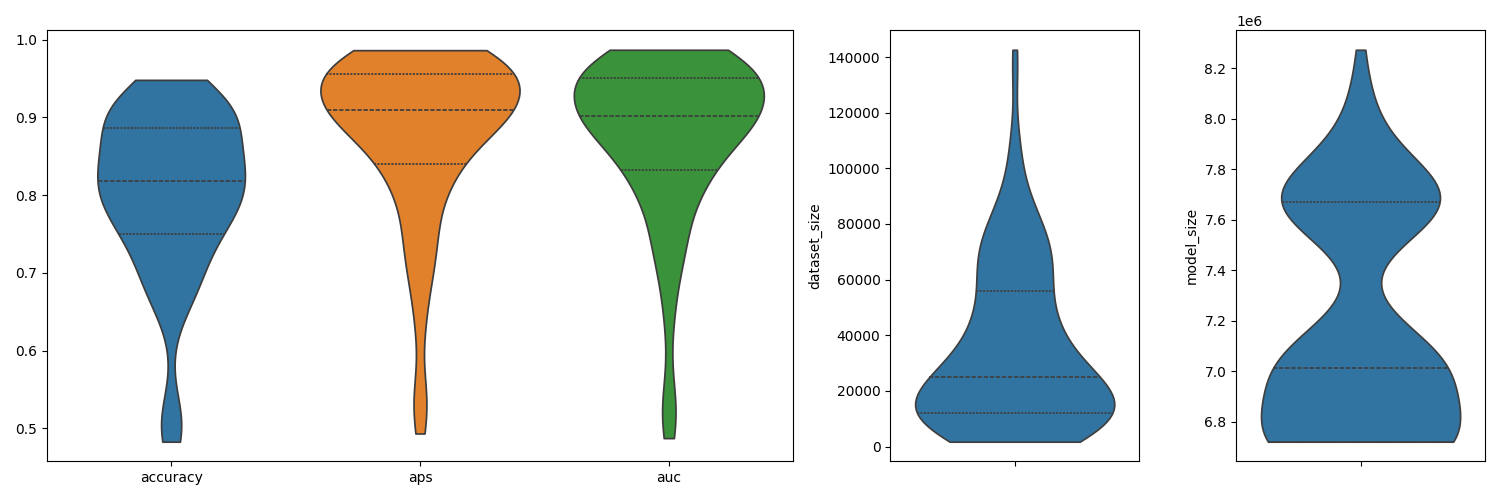}
        \caption{Violin plot of AUC, APs and accuracy (left) dataset size (middle) and model size in terms
        of \# of parameters (right) across the 165 binary datasets}
        \label{fig:bin_alltogheter}
    \end{figure}

\begin{figure}
        \centering
        \includegraphics[width=1\linewidth]{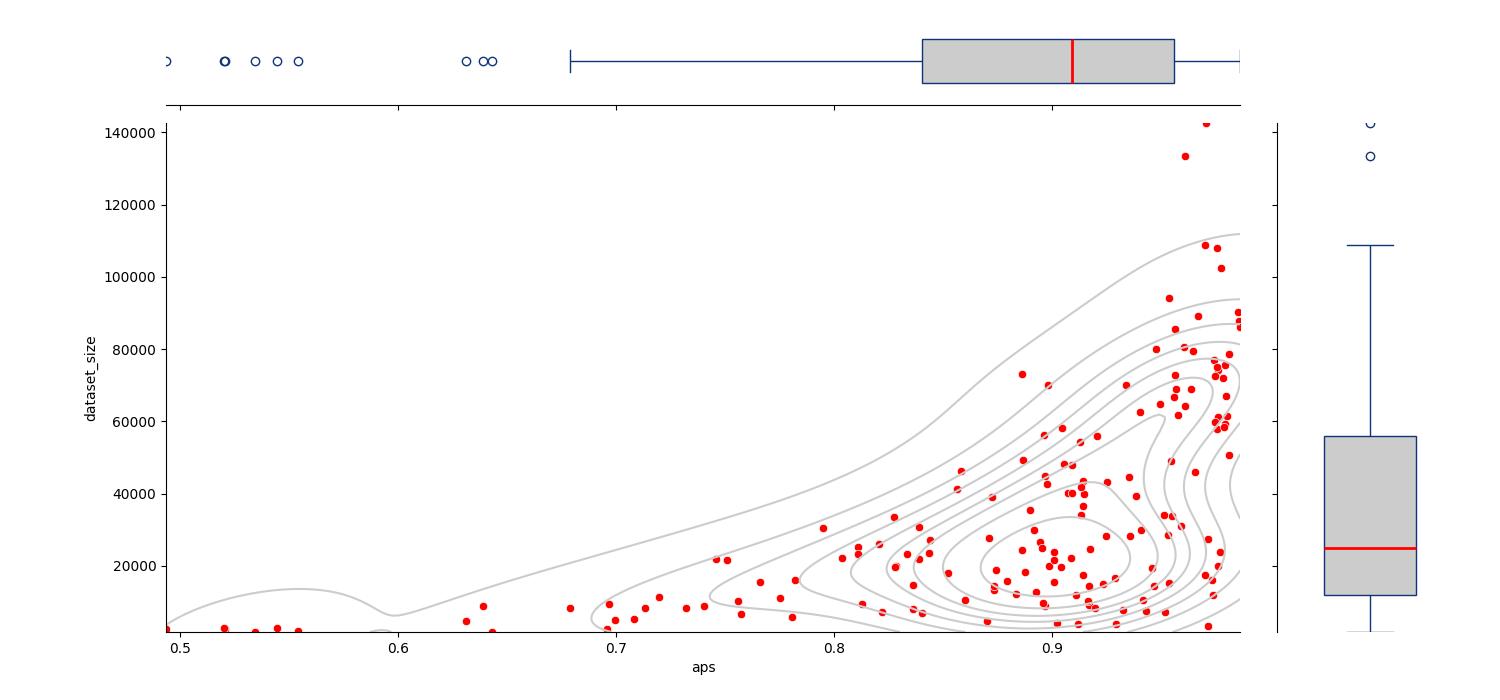}
        \caption{Joint distribution plot of AP and dataset size. Marginal plots show as boxplots the marginal distributions
        of AP (x-axis) and dataset size (y-axis). The joint plot shows the bi-variate distribution as a scatterplot
        annotated by contour lines computed with Kernel Density Estimate (KDE)}
        \label{fig:bin_dist}
    \end{figure}

    \noindent Overall, the TCN-based model adapted to binary classification achieved satisfactory performances detailed in
    Table~\ref{tab:binsum} and Fig. \ref{fig:bin_alltogheter}, comparable and in line with the
    state-of-the-art of TFBS binary classification~\cite{zhang2024mlsnet, ding2023deepstf}.
    The comparison is unfavorable for our model as we are comparing it with deep learning models specifically developed
    for binary classification, some of which are also \textit{trained with DNA shape data} to further improve on the task
    and thus having access to more data modalities besides sequence features.
    It is important to note, as highlighted by Figure~\ref{fig:bin_dist}, that while there is a moderate correlation
    between metrics and dataset size, namely a Pearson coefficient of $0.61$, $0.56$ and $0.57$ for accuracy, AP and AU-ROC respectively,
    the TCN-based model obtained outstanding results also on small datasets,
    achieving less than $0.7$ AP only on a 13 small size datasets out of 165.
    These results confirm the suitability of our architecture for TFBS classification tasks in general, moreover the model
    exhibited robust performances also on several small-sized datasets, showing a satisfactory behaviour also while
    trained in a regime of data scarcity.

 %   \begin{figure}
 %       \centering
 %       \includegraphics[width=1\linewidth]{aux_files/graphs/model_size_aps}
 %       \caption{Joint distribution plot of APs and model size \textcolor{red}{Personalmente la rimuoverei}}
 %       \label{fig:model_size_aps}
 %   \end{figure}

    \subsection{Multi-label classification results}
    We trained and tested our TCN-based model as well as the Bi-LSTM-based baseline on the three multi label dataset.
    We evaluated the performance of the trained models using both label-specific and summary metrics.
    In order to assess the label-specific metrics we used F1-score~\cite{christen2023review}, precision and recall~\cite{powers2020evaluation}.
    As summary metrics we adopted Area Under the Receiving-Operator Curve (AUC or AU-ROC)~\cite{christen2023review} and Average Precision (AP) defined as:
    \[
        AP = \sum_n (R_n-R_{n-1})P_n
    \] where $R_n, P_n$ are respectively precision and recall at the $n^{th}$ decision threshold.
    AP does approximate the area under the Precision-Recall curve without using linear interpolation.
    It has been noted that estimating the area under the curve with linear interpolation results in an overly optimistic metric~\cite{davis2006relationship, flach2015precision},
    while AP is a more conservative approximation.
    In addition, it is important to note that, while it is a widespread and generally accepted metric, AU-ROC is not suited to compare performances across different datasets.
    This is due to the fact that the baseline value depends on the dataset composition, making comparison across differently skewed datasets rather challenging.
    This is particularly relevant in cases where the class imbalance is severe as it is in ours. \\

\begin{figure}[h!]
        \centering
        \includegraphics[width=1\linewidth]{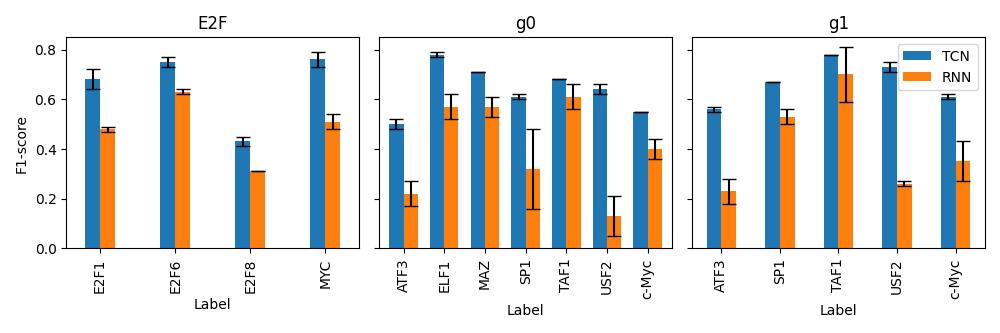}
        \caption{F1-score comparison across the 3 different datasets. }
        \label{fig:barchart}
    \end{figure}

    \noindent Overall, the TCN-based model outperformed the baseline on almost all metrics across all datasets and labels.
    The TCN-based model obtained a general significant gain over the RNN-based model both in terms of performance and stability as can be seen from the plot of F1 scores in Figure \ref{fig:barchart}.
    We will now discuss in details the performance of both models on each joint dataset.

    \par{\textbf{The H-M-E2F dataset.}}
    The H-M-E2F datasets represents the less complex datasets in terms of number of labels and the smallest in terms of number of sequences.
    The TCN-based model demonstrated a clear and substantial gain over the baseline, achieving satisfactory performance in areas where the baseline model proved inadequate.
    In terms of AP and AUC, the TCN-model achieved a stable and considerable gain.
    On the labels specific metrics, the highest gain for F1 and recall was obtained on the most frequent class, namely \textit{MYC}, this improvement is also reflected in the gain obtained in the samples average which shows the highest increase over the considered averaging methods as expected.
    The class with the highest precision improvement on the other hand is \textit{E2F1}.
    The performance on \textit{E2F8}, while showing a similar gain in magnitude compared to the other labels, remain underwhelming and significantly lower than performance obtained on the other labels.

    \begin{table}[ht]
\centering
\resizebox{\textwidth}{!}{
\begin{tabular}{l|rrr|rrr|rrr|r}
\hline
\textbf{Model} & \multicolumn{3}{c}{\textbf{TCN}} & \multicolumn{3}{c}{\textbf{RNN}} & \multicolumn{3}{c}{\textbf{$\Delta_{TCN - RNN}$}} \\
\hline
\textbf{Label} & f1-score & precision & recall & f1-score & precision & recall & f1-score & precision & recall & support \\
\hline
E2F1 & $0.68\pm0.04$ & $0.69\pm0.09$ & $0.70\pm0.14$ & $0.48\pm0.01$ & $0.39\pm0.00$ & $0.62\pm0.03$ & $+0.20$ & \cellcolor[gray]{0.8}\textcolor{red}{$+0.30$} & $+0.08$ & 3876 \\
E2F6 & $0.75\pm0.02$ & $0.79\pm0.04$ & $0.72\pm0.07$ & $0.63\pm0.01$ & $0.66\pm0.02$ & $0.61\pm0.03$ & $+0.12$ & $+0.13$ & $+0.11$ & 4338 \\
E2F8 & $0.43\pm0.02$ & $0.34\pm0.03$ & $0.58\pm0.09$ & $0.31\pm0.00$ & $0.24\pm0.00$ & $0.45\pm0.03$ & $+0.12$ & $+0.10$ & $+0.13$ & 1977 \\
MYC  & $0.76\pm0.03$ & $0.80\pm0.02$ & $0.73\pm0.06$ & $0.51\pm0.03$ & $0.71\pm0.00$ & $0.40\pm0.04$ & \cellcolor[gray]{0.8}\textcolor{red}{$+0.25$} & $+0.09$ & \cellcolor[gray]{0.8}\textcolor{red}{$+0.33$} & 7012 \\
\hline
macro avg & $0.65\pm0.02$ & $0.66\pm0.03$ & $0.68\pm0.06$ & $0.48\pm0.01$ & $0.50\pm0.00$ & $0.52\pm0.02$ & $+0.17$ & $+0.16$ & $+0.16$ & 17203 \\
micro avg & $0.69\pm0.01$ & $0.68\pm0.04$ & $0.70\pm0.04$ & $0.50\pm0.01$ & $0.49\pm0.00$ & $0.51\pm0.02$ & $+0.19$ & $+0.19$ & $+0.19$ &  \\
samples avg & $0.68\pm0.02$ & $0.72\pm0.04$ & $0.72\pm0.03$ & $0.42\pm0.01$ & $0.45\pm0.01$ & $0.48\pm0.02$ & $+0.26$ & $+0.27$ & $+0.24$ &  \\
weighted avg & $0.70\pm0.01$ & $0.72\pm0.03$ & $0.70\pm0.04$ & $0.51\pm0.01$ & $0.57\pm0.00$ & $0.51\pm0.02$ & $+0.19$ & $+0.15$ & $+0.19$ &  \\
\hline
\multicolumn{11}{c}{\textbf{Summary Metrics}} \\
\hline
APS & $0.73\pm0.01$ & & & $0.52\pm0.00$ & & & $+0.21$ & & & \\
AUC & $0.80\pm0.01$ & & & $0.59\pm0.00$ & & & $+0.21$ & & & \\
\hline
\end{tabular}
}
\caption{Performance of the implemented models on E2F dataset. The reported metrics are averaged across 5 runs and are reported alongside standard deviation. The rightmost part of the table shows the gain of the TCN-based model over the baseline. Highest gain for each metric is highligthed in red.}
\end{table}

    \par{\textbf{The \gzero dataset.}}
    The \gzero dataset is the biggest dataset taken into consideration, both in terms of number of labels and training sequences.
    The TCN-based model achieved also on this dataset a substantial gain over the baseline model, achieving higher gains compared to H-M-E2F dataset.
    It is important to note, however, that the higher gains are largely due to the fact that the baseline performance is way lower compared to the one achieved on the H-M-E2F dataset.
    This is particularly evident by comparing the AP scores of the two models on the two datasets; the TCN-based model's AP score are somewhat comparable while the baseline model clearly
    struggles more on \gzero.\\
    Taking into consideration label specific metrics, it is particularly noteworthy that the highest gain for each metric has been obtained on the same label, \textbf{USF2}.
    This is particularly interesting because USF2 is the least frequent class in \gzero, this suggests that the TCN-based model's gain are not just imputable to an improved overall capacity to leverage training data and that the TCN-based model is able to capture and learn label-specific features that the recurrent baseline cannot, even with fewer examples.
    This in turns suggests that USF2 label is characterized by different sequence features that a recurrent architecture cannot fully learn.
    It is also worth noting that the performance of the TCN-based model are more stable compared to the baseline as can bee seen from the metrics' standard deviations.

    \begin{table}[ht]
\centering
\resizebox{\textwidth}{!}{
\begin{tabular}{l|rrr|rrr|rrr|r}
\hline
\textbf{Model} & \multicolumn{3}{c}{\textbf{TCN}} & \multicolumn{3}{c}{\textbf{RNN}} & \multicolumn{3}{c}{\textbf{$\Delta_{TCN - RNN}$}} \\
\hline
\textbf{Label} & f1-score & precision & recall & f1-score & precision & recall & f1-score & precision & recall & support \\
\hline
ATF3 & $0.50\pm0.02$ & $0.45\pm0.08$ & $0.59\pm0.13$ & $0.22\pm0.01$ & $0.15\pm0.01$ & $0.45\pm0.20$ & $+0.28$ & $+0.30$ & $+0.14$ & 3257 \\
ELF1 & $0.78\pm0.01$ & $0.86\pm0.02$ & $0.71\pm0.03$ & $0.57\pm0.05$ & $0.59\pm0.05$ & $0.57\pm0.14$ & $+0.21$ & $+0.27$ & $+0.14$ & 9660 \\
MAZ & $0.71\pm0.00$ & $0.65\pm0.01$ & $0.78\pm0.02$ & $0.57\pm0.04$ & $0.58\pm0.05$ & $0.58\pm0.14$ & $+0.14$ & $+0.07$ & $+0.20$ & 9350 \\
SP1 & $0.61\pm0.01$ & $0.53\pm0.01$ & $0.71\pm0.03$ & $0.32\pm0.16$ & $0.36\pm0.07$ & $0.41\pm0.26$ & $+0.29$ & $+0.17$ & $+0.30$ & 7200 \\
TAF1 & $0.68\pm0.00$ & $0.61\pm0.01$ & $0.76\pm0.01$ & $0.61\pm0.05$ & $0.51\pm0.10$ & $0.78\pm0.09$ & $+0.07$ & $+0.10$ & $-0.02$ & 5859 \\
USF2 & $0.64\pm0.02$ & $0.54\pm0.05$ & $0.80\pm0.04$ & $0.13\pm0.08$ & $0.09\pm0.05$ & $0.28\pm0.19$ & \cellcolor[gray]{0.8}\textcolor{red}{$+0.51$} & \cellcolor[gray]{0.8}\textcolor{red}{$+0.45$} & \cellcolor[gray]{0.8}\textcolor{red}{$+0.52$} & 2460 \\
c-Myc & $0.55\pm0.00$ & $0.47\pm0.01$ & $0.66\pm0.02$ & $0.40\pm0.04$ & $0.34\pm0.01$ & $0.50\pm0.15$ & $+0.15$ & $+0.13$ & $+0.16$ & 6891 \\
\hline
macro avg & $0.64\pm0.01$ & $0.59\pm0.02$ & $0.72\pm0.03$ & $0.40\pm0.02$ & $0.37\pm0.03$ & $0.51\pm0.05$ & $+0.24$ & $+0.22$ & $+0.21$ & 44677 \\
micro avg & $0.65\pm0.01$ & $0.60\pm0.01$ & $0.72\pm0.02$ & $0.44\pm0.03$ & $0.38\pm0.02$ & $0.54\pm0.07$ & $+0.21$ & $+0.22$ & $+0.18$ &  \\
samples avg & $0.64\pm0.01$ & $0.63\pm0.01$ & $0.75\pm0.01$ & $0.33\pm0.05$ & $0.30\pm0.05$ & $0.49\pm0.09$ & $+0.31$ & $+0.33$ & $+0.26$ &  \\
weighted avg & $0.66\pm0.01$ & $0.62\pm0.01$ & $0.72\pm0.02$ & $0.46\pm0.02$ & $0.44\pm0.03$ & $0.54\pm0.07$ & $+0.20$ & $+0.18$ & $+0.18$ &  \\
\hline
\multicolumn{11}{c}{\textbf{Summary Metrics}} \\
\hline
APS & $0.69\pm0.01$ & & & $0.37\pm0.00$ & & & $+0.32$ & & & \\
AUC & $0.84\pm0.00$ & & & $0.60\pm0.01$ & & & $+0.24$ & & & \\
\hline
\end{tabular}
}
\caption{Performance of the implemented models on g0 dataset. The reported metrics are averaged across 5 runs and are reported alongside standard deviation. The rightmost part of the table shows the gain of the TCN-based model over the baseline. Highest gain for each metric is highligthed in red.}
\end{table}

    \par{\bf The \gone dataset.}
    The \gone dataset lies in a middle ground between H-M-E2F and \gzero dataset, both in terms of labels and in terms of training samples.
    In fact, it is composed by one more label compared to H-M-E2F while being constituted by almost twice the samples.
    Broadly speaking both models follow the same trend observed for \gzero with the TCN-based model outperforming the baseline.
    The AP scores of both models are comparable with the ones achieved on the \gzero dataset as well.
    As befeore, the highest gain has been achieved on the less frequent class, USF2, for both F1 and precision while the highest gain in terms of precision has been obtained on c-Myc.

   \begin{table}[ht]
\centering
\resizebox{\textwidth}{!}{
\begin{tabular}{l|rrr|rrr|rrr|r}
\hline
\textbf{Model} & \multicolumn{3}{c}{\textbf{TCN}} & \multicolumn{3}{c}{\textbf{RNN}} & \multicolumn{3}{c}{\textbf{Diff (TCN - RNN)}} \\
\hline
\textbf{Label} & f1-score & precision & recall & f1-score & precision & recall & f1-score & precision & recall & support \\
\hline
ATF3 & $0.56\pm0.01$ & $0.53\pm0.06$ & $0.61\pm0.08$ & $0.23\pm0.05$ & $0.18\pm0.01$ & $0.40\pm0.18$ & $+0.33$ & $+0.35$ & $+0.21$ & 3952 \\
SP1 & $0.67\pm0.00$ & $0.68\pm0.02$ & $0.65\pm0.02$ & $0.53\pm0.03$ & $0.45\pm0.01$ & $0.67\pm0.10$ & $+0.14$ & $+0.23$ & $-0.02$ & 9452 \\
TAF1 & $0.78\pm0.00$ & $0.77\pm0.04$ & $0.79\pm0.05$ & $0.70\pm0.11$ & $0.77\pm0.05$ & $0.67\pm0.18$ & $+0.08$ & $+0.00$ & $+0.12$ & 8615 \\
USF2 & $0.73\pm0.02$ & $0.69\pm0.06$ & $0.78\pm0.05$ & $0.26\pm0.01$ & $0.16\pm0.00$ & $0.67\pm0.11$ & \cellcolor[gray]{0.8}\textcolor{red}{$+0.47$} & \cellcolor[gray]{0.8}\textcolor{red}{$+0.53$} & $+0.11$ & 3457 \\
c-Myc & $0.61\pm0.01$ & $0.57\pm0.03$ & $0.65\pm0.04$ & $0.35\pm0.08$ & $0.37\pm0.01$ & $0.36\pm0.13$ & $+0.26$ & $+0.20$ & \cellcolor[gray]{0.8}\textcolor{red}{$+0.29$} & 7889 \\
\hline
macro avg & $0.67\pm0.01$ & $0.65\pm0.03$ & $0.70\pm0.03$ & $0.42\pm0.04$ & $0.38\pm0.01$ & $0.55\pm0.05$ & $+0.25$ & $+0.27$ & $+0.15$ & 33365 \\
micro avg & $0.67\pm0.00$ & $0.65\pm0.03$ & $0.70\pm0.02$ & $0.44\pm0.03$ & $0.36\pm0.01$ & $0.56\pm0.05$ & $+0.23$ & $+0.29$ & $+0.14$ &  \\
samples avg & $0.68\pm0.01$ & $0.70\pm0.02$ & $0.74\pm0.02$ & $0.41\pm0.02$ & $0.35\pm0.01$ & $0.58\pm0.05$ & $+0.27$ & $+0.35$ & $+0.16$ &  \\
weighted avg & $0.67\pm0.00$ & $0.66\pm0.02$ & $0.70\pm0.02$ & $0.47\pm0.04$ & $0.45\pm0.01$ & $0.56\pm0.05$ & $+0.20$ & $+0.21$ & $+0.14$ &  \\
\hline
\multicolumn{11}{c}{\textbf{Summary Metrics}} \\
\hline
APS & $0.73\pm0.01$ & & & $0.38\pm0.00$ & & & $+0.35$ & & & \\
AUC & $0.84\pm0.00$ & & & $0.58\pm0.01$ & & & $+0.26$ & & & \\
\hline
\end{tabular}
}
\caption{Performance of the implemented models on g1 dataset. The reported metrics are averaged across 5 runs and are reported alongside standard deviation. The rightmost part of the table shows the gain of the TCN-based model over the baseline. Highest gain for each metric is highligthed in red.}
\end{table}
    \begin{table}[ht]
    \centering
    \resizebox{\textwidth}{!}{
        \begin{tabular}{l|rrr|rrr|rrr}
            \hline
            \textbf{Label} & \multicolumn{3}{c}{\textbf{$\Delta$ E2F}} & \multicolumn{3}{c}{\textbf{$\Delta$ g0}} & \multicolumn{3}{c}{\textbf{$\Delta$ g1}} \\
            \hline
            & \textbf{f1-score} & \textbf{precision} & \textbf{recall} & \textbf{f1-score} & \textbf{precision} & \textbf{recall} & \textbf{f1-score} & \textbf{precision} & \textbf{recall} \\
            \hline
            \multicolumn{10}{c}{\textbf{Medie}} \\
            \hline
            macro avg     & $+0.17$ & $+0.16$ & $+0.16$ & $+0.24$ & $+0.22$ & $+0.21$ & $+0.25$ & $+0.27$ & $+0.15$ \\
            micro avg     & $+0.19$ & $+0.19$ & $+0.19$ & $+0.21$ & $+0.22$ & $+0.18$ & $+0.23$ & $+0.29$ & $+0.14$ \\
            samples avg   & $+0.26$ & $+0.27$ & $+0.24$ & $+0.31$ & $+0.33$ & $+0.26$ & $+0.27$ & $+0.35$ & $+0.16$ \\
            weighted avg  & $+0.19$ & $+0.15$ & $+0.19$ & $+0.20$ & $+0.18$ & $+0.18$ & $+0.20$ & $+0.21$ & $+0.14$ \\
            \hline
            \multicolumn{10}{c}{\textbf{Summary Metrics}} \\
            \hline
            APS           & $+0.21$ &           &           & $+0.32$ &           &           & $+0.35$ &           &           \\
            AUC           & $+0.21$ &           &           & $+0.24$ &           &           & $+0.26$ &           &           \\
            \hline
        \end{tabular}
    }
    \caption{Gain (TCN -- RNN) delle tre tabelle, fianco a fianco}\label{tab:table}
\end{table}
%\newpage
    \subsection{Attributions}
    In order to understand what the models' have been learning and to gain insight on the
    effectiveness of our trained TCN-based model, we applied explainability techniques.
    We derived attribution scores for the TCN-based model trained on the H-M-E2F dataset with Integrated Gradients and used
    TF‑MoDISco \cite{shrikumar2018technical} to identify the most informative seqlets.

    \begin{figure}[h!]
        \centering
        \begin{subfigure}{0.75\textwidth}

            \includegraphics[width=1\linewidth]{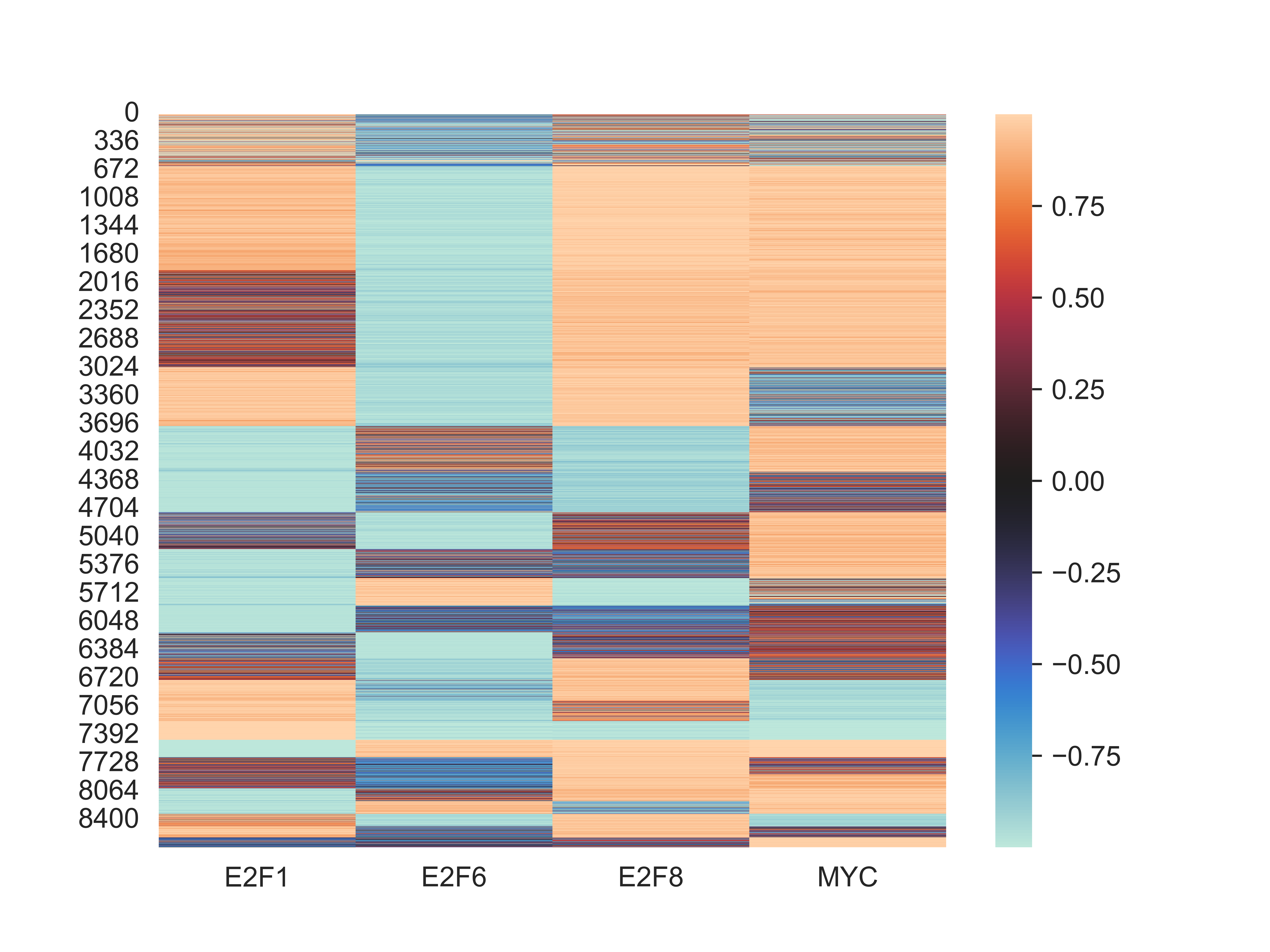}
            \caption{Activity heatmap of all the relevant seq-lets (x-axis) identified by MoDisco from the attribution scores.
            Each seqlet is associated to an activity pattern across the labes (y-axis).
            Positive values show increased affinity for the label, i.e. the presence of the seqlet postivively inflences
            the prediction of the model thowards the correspondign label, and conversely negative value show negative
            influence of the seqlet on the prediction of the associated label.
            It is worth noting that several seqlets exibit similar activity patterns across all labels suggesting
            an underlying biological mechanism.
            Additional information can be found in the MoDisco documentation.}
            \label{subfig:heat}
        \end{subfigure}

        \begin{subfigure}{0.75\textwidth}
        \centering
        \includegraphics[width=1\linewidth]{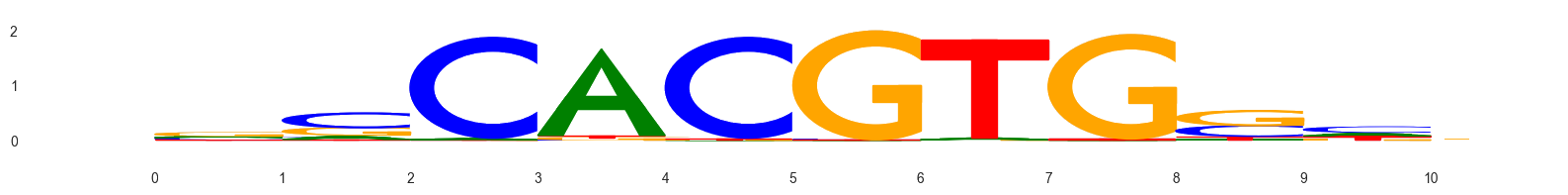}
        \caption{motif logos of identified seq let corresponding to MYC conensus sequence}
        \label{subfig:MYClogo}

        \end{subfigure}
        \centering
        \begin{subfigure}{0.75\textwidth}
        \includegraphics[width=1\linewidth]{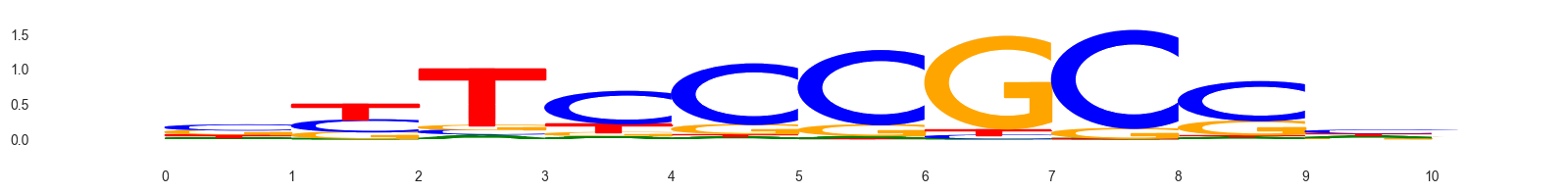}
        \caption{motif logos of identified seq let corresponding to E2F6 conensus sequence}
        \label{subfig:E2Flogo}
        \end{subfigure}

        \caption{
        Preliminary result of the attribution pipeline}

        \label{fig:attr}

    \end{figure}

    The heat-map and sequence logos obtained, as in Figure~\ref{fig:attr},
    clearly show that the model is capturing correctly at least part of the
    underlying biological mechanism.
    In fact, the box logos obtained from the attribution pipeline represent two well known motifs belonging to consensus
    sequences of the labels present in the training dataset, namely MYC and E2F6.

    The nature of the proposed task at hand does not allow for a seamless integration of existing attribution pipelines, due to the multi-label nature of our task it would be preferable to modify and apply attribution methods accordingly.
    The proper development of a sound attribution pipeline devised specifically for multi-label data, however, is out of the scope of this study and will be considered in future works.

    \section{Conclusions}

We address the question of multiple Transcription Factors (TF)
DNA-binding recognition 
modelled via multiple labels, 
going beyond the binary, single-TF prediction paradigm.
%By jointly modeling the binding of multiple TFs directly from DNA sequence, we aimed to capture aspects of the combinatorial and cooperative nature of transcriptional regulation that are difficult to observe when TFs are considered in isolation.
Our deep learning framework based on Temporal Convolutional Networks (TCNs)
achieves an effective learning in settings characterized by limited and noisy biological data,
with a significant predictive performance.
Beyond the predictive accuracy, we applied explainable artificial intelligence methods
to extract from our models biologically meaningful insights as
sequence motifs.
%\textcolor{red}{\sout{and co-binding patterns consistent with known single/multiple
%    transcription factor motifs and their cooperative interactions.} secondo me anche qui bisogna abbassare un po' il
%    tiro perche biologicamente sono affermazioni molto forti per cui servono risultati molto più solidi e analisi diverse.}
Our findings suggest that multi-label deep learning models for
%\textcolor{red}{\sout{TF binding site}
{TFBS} prediction can serve not just as a predictive tool, but also as an hypothesis-generating
framework for studying transcriptional regulation.
We plan to deepen our investigation in the future towards a
more comprehensive understanding of gene regulatory networks and their underlying
cooperative mechanisms.
%\textcolor{red}{Aggiungerei anche qualcosina sull'idea di sviluppare tecniche di attribuzioni
%cooerenti col framework multilabel per esempio qualcosa tipo: --
Furthermore, we plan to develop an attribution
framework specifically tailored on multi-label setting to fully leverage all the
information and insights learned by the trained models.

    \section{Data availability}

   All the experimental data used to construct our dataset is publicly available on
   ENCODE Consortium~\cite{encode2012integrated}.
    In order to construct the two data-driven datasets we downloaded all available Chip-Seq experiments for the TF
    and cell-lines listed in Table~\ref{tab:g}.
    For the manually curated dataset we downloaded Chip-Seq targeting the already specified transcription factor from
    in K562 cell line from
    %  the ENCODE Regulation 'TF Clusters'
    % For the manually curated dataset we downloaded Chip-Seq targeting E2F1, E2F6, E2F8, MYC in K562 cell line from
   %  the ENCODE Regulation 'TF Clusters' track.\textcolor{red}{F. Va scritto prima}
    The dataset used to test the capability of our model in the binary setting is the dataset proposed by~\cite{zeng2016convolutional}.
    The full list of encode identifiers is available upon request.

    \section{Code availability}
    The code is fully available upon request to the authors.

    \section{Conflict of interest}
    All authors declare no conflict of interest.

% bibliography - bibtex format
%
% add chapter to index
    \addcontentsline{toc}{section}{References}
% alphabetical order of authors
    \bibliographystyle{plain}
    \bibliography{aux_files/references_clean}

@article{ahmadi2021myc,
  author    = {Ahmadi, Seyed Esmaeil and Rahimi, Samira and Zarandi, Bahman and Chegeni, Rouzbeh and Safa, Majid},
  title     = {MYC: a multipurpose oncogene with prognostic and therapeutic implications in blood malignancies},
  journal   = {Journal of Hematology \& Oncology},
  year      = {2021},
  volume    = {14},
  number    = {1},
  pages     = {121},
  doi       = {10.1186/s13045-021-01111-4},
  url       = {https://link.springer.com/article/10.1186/s13045-021-01111-4}
}

@article{alon2007network,
  author = {Alon, Uri},
  title = {Network motifs: theory and experimental approaches},
  journal = {Nature Reviews Genetics},
  year = {2007},
  volume = {8},
  number = {6},
  pages = {450--461}
}

@misc{bai2018empirical,
  author    = {Bai, Shaojie and Kolter, J. Zico and Koltun, Vladlen},
  title     = {An Empirical Evaluation of Generic Convolutional and Recurrent Networks for Sequence Modeling},
  year      = {2018},
  archivePrefix = {arXiv},
  eprint    = {1803.01271},
  primaryClass = {cs.LG},
  doi       = {10.48550/arXiv.1803.01271},
  url       = {https://arxiv.org/abs/1803.01271}
}

@article{christen2023review,
  author    = {Christen, Peter and Hand, David J and Kirielle, Nishadi},
  title     = {A Review of the F-Measure: Its History, Properties, Criticism, and Alternatives},
  journal   = {ACM Computing Surveys},
  year      = {2024},
  volume    = {56},
  number    = {3},
  pages     = {73:1--73:24},
  doi       = {10.1145/3606367}
}

@article{encode2012integrated,
  author = {ENCODE Project Consortium and others},
  title = {An integrated encyclopedia of DNA elements in the human genome},
  journal = {Nature},
  year = {2012},
  volume = {489},
  number = {7414},
  pages = {57--74},
  doi    = {10.1038/nature11247}
}

@article{hochreiter1997long,
  title={Long short-term memory},
  author={Hochreiter, Sepp and Schmidhuber, J{"u}rgen},
  journal={Neural computation},
  volume={9},
  number={8},
  pages={1735--1780},
  year={1997},
  publisher={MIT press}
}

@misc{kondratyuk2021movinets,
  author = {Dan Kondratyuk and Liangzhe Yuan and Yandong Li and Li Zhang and Mingxing Tan and Matthew Brown and Boqing Gong},
  title = {MoViNets: Mobile Video Networks for Efficient Video Recognition},
  year = {2021},
  archiveprefix = {arXiv},
  eprint = {2103.11511},
  primaryclass = {cs.CV}
}

@article{lara-benitez_temporal_2020,
  author = {Lara-Ben{\' i}tez, Pedro and Carranza-Garc{\' i}a, Manuel and Luna-Romera, Jos{\' e} M. and Riquelme, Jos{\' e} C.},
  title = {Temporal {Convolutional} {Networks} {Applied} to {Energy}-{Related} {Time} {Series} {Forecasting}},
  journal = {Applied Sciences},
  year = {2020},
  volume = {10},
  number = {7},
  pages = {2322},
  doi = {10.3390/app10072322},
  url = {https://www.mdpi.com/2076-3417/10/7/2322},
  issn = {2076-3417}
}

@misc{reback2020pandas,
  author = {The pandas development team},
  title = {pandas-dev/pandas: Pandas},
  year = {2020},
  publisher = {Zenodo},
  doi = {10.5281/zenodo.3509134},
  url = {https://doi.org/10.5281/zenodo.3509134},
  version = {latest}
}

@article{reiter2017direct,
  author = {Reiter, Franziska and Wienerroither, Sebastian and Stark, Alexander},
  title  = {Combinatorial function of transcription factors and cofactors},
  journal = {Current Opinion in Genetics \& Development},
  year   = {2017},
  volume = {43},
  pages  = {73--81},
  doi    = {10.1016/j.gde.2016.12.007}
}

@article{spitz2012transcription,
  author = {Spitz, Fran{\c c}ois and Furlong, Eileen E. M.},
  title = {Transcription factors: from enhancer binding to developmental control},
  journal = {Nature Reviews Genetics},
  year = {2012},
  volume = {13},
  number = {9},
  pages = {613--626},
  doi    = {10.1038/nrg3207}
}

@inproceedings{sundararajan2017axiomatic,
  author = {Sundararajan, Mukund and Taly, Ankur and Yan, Qiqi},
  title = {Axiomatic attribution for deep networks},
  booktitle = {International conference on machine learning},
  year = {2017},
  pages = {3319--3328},
  organization = {PMLR}
}

@article{zeng2016convolutional,
  author = {Zeng, Haoyang and Edwards, Matthew D and Liu, Ge and Gifford, David K},
  journal = {Bioinformatics},
  year = {2016},
  volume = {32},
  number = {12},
  pages = {i121--i127},
  publisher = {Oxford University Press}
}

@article{amati1992myc,
  author = {Amati, Bruno and Dalton, Stephen and Brooks, Michael W. and Littlewood, Thomas D. and Evan, Gerard I. and Land, Hartmut},
  title = {Myc and Max form a sequence-specific DNA-binding protein complex},
  journal = {Nature},
  year = {1992},
  volume = {359},
  pages = {423--426}
}

@article{ding2023deepstf,
  author = {Ding, Pengju and Wang, Yifei and Zhang, Xinyu and Gao, Xin and Liu, Guozhu and Yu, Bin},
  title = {DeepSTF: predicting transcription factor binding sites by interpretable deep neural networks combining sequence and shape},
  journal = {Briefings in bioinformatics},
  year = {2023},
  volume = {24},
  number = {4},
  pages = {bbad231},
  publisher = {Oxford University Press}
}

@article{fioresi2022deep,
  author = {Fioresi, Rita and Demurtas, P and Perini, G},
  title = {Deep Learning for MYC binding site recognition},
  journal = {Frontiers in Bioinformatics},
  year = {2022},
  volume = {2},
  pages = {1015993},
  publisher = {Frontiers}
}

@article{han2019,
  author = {Yuan, Han and Kshirsagar, Meghana and Zamparo, Lee and Lu, Yuheng and Leslie, Christina},
  title = {BindSpace decodes transcription factor binding signals by large-scale sequence embedding},
  journal = {Nature Methods},
  year = {2019},
  volume = {16},
  pages = {1--4},
  doi    = {10.1038/s41592-019-0511-y}
}

@article{latchman1997transcription,
  author = {Latchman, David S},
  title = {Transcription factors: an overview},
  journal = {The international journal of biochemistry \& cell biology},
  year = {1997},
  volume = {29},
  number = {12},
  pages = {1305--1312},
  publisher = {Elsevier}
}

@article{morgunova2017structural,
  author = {Morgunova, Ekaterina and Taipale, Jussi},
  title = {Structural perspective of cooperative transcription factor binding},
  journal = {Current opinion in structural biology},
  year = {2017},
  volume = {47},
  pages = {1--8},
  publisher = {Elsevier}
}

@article{shrikumar2018technical,
  author = {Shrikumar, Avanti and Tian, Katherine and Avsec, {\v Z}iga and Shcherbina, Anna and Banerjee, Abhimanyu and Sharmin, Mahfuza and Nair, Surag and Kundaje, Anshul},
  title = {Technical note on transcription factor motif discovery from importance scores (TF-MoDISco) version 0.5. 6.5},
  journal = {arXiv preprint arXiv:1811.00416},
  year = {2018}
}

@article{Xie2025DNAguidedTF,
  author    = {Xie, Zhiyuan and Sokolov, Ilya and Osmala, Maria and Yue, Xue and Bower, Grace and Pett, Jan Patrick and Chen, Yinan and Wang, Kai and Cavga, Ayse Derya and Popov, Alexander and Teichmann, Sarah A. and Morgunova, Ekaterina and Kvon, Evgeny Z. and Yin, Yimeng and Taipale, Jussi},
  title     = {DNA-guided transcription factor interactions extend human gene regulatory code},
  journal   = {Nature},
  year      = {2025},
  volume    = {641},
  pages     = {1329--1338},
  doi       = {10.1038/s41586-025-08844-z},
  url       = {https://helda.helsinki.fi/bitstreams/eb776a9b-fb70-47e7-81ad-df341d38bce0/download}
}

@article{zhang2022novel,
  author = {Zhang, Yongqing and Wang, Zixuan and Zeng, Yuanqi and Liu, Yuhang and Xiong, Shuwen and Wang, Maocheng and Zhou, Jiliu and Zou, Quan},
  title = {A novel convolution attention model for predicting transcription factor binding sites by combination of sequence and shape},
  journal = {Briefings in Bioinformatics},
  year = {2022},
  volume = {23},
  number = {1},
  publisher = {Oxford Academic}
}

@article{zhang2024mlsnet,
  title={MLSNet: a deep learning model for predicting transcription factor binding sites},
  author={Zhang, Yuchuan and Wang, Zhikang and Ge, Fang and Wang, Xiaoyu and Zhang, Yiwen and Li, Shanshan and Guo, Yuming and Song, Jiangning and Yu, Dong-Jun},
  journal={Briefings in Bioinformatics},
  volume={25},
  number={6},
  pages={bbae489},
  year={2024},
  publisher={Oxford University Press}
}

@inproceedings{ansel2024pytorch,
  author = {Ansel, Jason and Yang, Edward and He, Horace and Gimelshein, Natalia and Jain, Animesh and Voznesensky, Michael and Bao, Bin and Bell, Peter and Berard, David and Burovski, Evgeni and others},
  title = {Pytorch 2: Faster machine learning through dynamic python bytecode transformation and graph compilation},
  booktitle = {Proceedings of the 29th ACM International Conference on Architectural Support for Programming Languages and Operating Systems, Volume 2},
  year = {2024},
  pages = {929--947}
}

@article{bengio2000neural,
  author = {Bengio, Yoshua and Ducharme, R{\' e}jean and Vincent, Pascal},
  title = {A neural probabilistic language model},
  journal = {Advances in neural information processing systems},
  year = {2000},
  volume = {13}
}

@article{bergstra2011algorithms,
  author = {Bergstra, James and Bardenet, R{\' e}mi and Bengio, Yoshua and K{\' e}gl, Bal{\' a}zs},
  title = {Algorithms for hyper-parameter optimization},
  journal = {Advances in neural information processing systems},
  year = {2011},
  volume = {24}
}

@inproceedings{bergstra2013making,
  author = {Bergstra, James and Yamins, Daniel and Cox, David},
  title = {Making a science of model search: Hyperparameter optimization in hundreds of dimensions for vision architectures},
  booktitle = {International conference on machine learning},
  year = {2013},
  pages = {115--123},
  organization = {PMLR}
}

@article{brauwers2021general,
  author = {Brauwers, Gianni and Frasincar, Flavius},
  title = {A general survey on attention mechanisms in deep learning},
  journal = {IEEE Transactions on Knowledge and Data Engineering},
  year = {2021},
  volume = {35},
  number = {4},
  pages = {3279--3298},
  publisher = {IEEE}
}

@article{buchler2003combinatorial,
  author = {Buchler, Nicolas E. and Gerland, Ulrich and Hwa, Terence},
  title = {On schemes of combinatorial transcription logic},
  journal = {Proceedings of the National Academy of Sciences},
  year = {2003},
  volume = {100},
  number = {9},
  pages = {5136--5141}
}

@inproceedings{davis2006relationship,
  author = {Davis, Jesse and Goadrich, Mark},
  title = {The relationship between Precision-Recall and ROC curves},
  booktitle = {Proceedings of the 23rd international conference on Machine learning},
  year = {2006},
  pages = {233--240}
}

@inproceedings{Devlin2019BERTPO,
  author = {Jacob Devlin and Ming-Wei Chang and Kenton Lee and Kristina Toutanova},
  title = {BERT: Pre-training of Deep Bidirectional Transformers for Language Understanding},
  booktitle = {North American Chapter of the Association for Computational Linguistics},
  year = {2019},
  url = {https://api.semanticscholar.org/CorpusID:52967399}
}

@article{flach2015precision,
  author = {Flach, Peter and Kull, Meelis},
  title = {Precision-recall-gain curves: PR analysis done right},
  journal = {Advances in neural information processing systems},
  year = {2015},
  volume = {28}
}

@article{harris2020array,
  title = {Array programming with {NumPy}},
  journal = {Nature},
  year = {2020},
  volume = {585},
  number = {7825},
  pages = {357--362},
  publisher = {Springer Science and Business Media {LLC}},
  doi = {10.1038/s41586-020-2649-2},
  url = {https://doi.org/10.1038/s41586-020-2649-2}
}

@inproceedings{he2016deep,
  author = {He, Kaiming and Zhang, Xiangyu and Ren, Shaoqing and Sun, Jian},
  title = {Deep residual learning for image recognition},
  booktitle = {Proceedings of the IEEE conference on computer vision and pattern recognition},
  year = {2016},
  pages = {770--778}
}

@article{powers2020evaluation,
  author = {Powers, David MW},
  title = {Evaluation: from precision, recall and F-measure to ROC, informedness, markedness and correlation},
  journal = {arXiv preprint arXiv:2010.16061},
  year = {2020}
}

@article{scikit-learn,
  author = {Pedregosa, F. and Varoquaux, G. and Gramfort, A. and Michel, V. and Thirion, B. and Grisel, O. and Blondel, M. and Prettenhofer, P. and Weiss, R. and Dubourg, V. and Vanderplas, J. and Passos, A. and Cournapeau, D. and Brucher, M. and Perrot, M. and Duchesnay, E.},
  title = {Scikit-learn: Machine Learning in {P}ython},
  journal = {Journal of Machine Learning Research},
  year = {2011},
  volume = {12},
  pages = {2825--2830}
}

@article{vaswani2017attention,
  author = {Vaswani, Ashish and Shazeer, Noam and Parmar, Niki and Uszkoreit, Jakob and Jones, Llion and Gomez, Aidan N and Kaiser, {\L}ukasz and Polosukhin, Illia},
  title = {Attention is all you need},
  journal = {Advances in neural information processing systems},
  year = {2017},
  volume = {30}
}

@article{zaharia2018accelerating,
  author = {Zaharia, Matei and Chen, Andrew and Davidson, Aaron and Ghodsi, Ali and Hong, Sue Ann and Konwinski, Andy and Murching, Siddharth and Nykodym, Tomas and Ogilvie, Paul and Parkhe, Mani and others},
  title = {Accelerating the machine learning lifecycle with MLflow.},
  journal = {IEEE Data Eng. Bull.},
  year = {2018},
  volume = {41},
  number = {4},
  pages = {39--45}
}

@article{bednarski_temporal_2022,
  author = {Bednarski, Bryan P. and Singh, Akash Deep and Zhang, Wenhao and Jones, William M. and Naeim, Arash and Ramezani, Ramin},
  title = {Temporal convolutional networks and data rebalancing for clinical length of stay and mortality prediction},
  journal = {Scientific Reports},
  year = {2022},
  volume = {12},
  number = {1},
  pages = {21247},
  doi = {10.1038/s41598-022-25472-z},
  url = {https://www.nature.com/articles/s41598-022-25472-z},
  issn = {2045-2322}
}

@misc{lea2016temporal,
  author = {Colin Lea and Rene Vidal and Austin Reiter and Gregory D. Hager},
  title = {Temporal Convolutional Networks: A Unified Approach to Action Segmentation},
  year = {2016},
  archiveprefix = {arXiv},
  eprint = {1608.08242},
  primaryclass = {cs.CV}
}

@misc{oord2016wavenet,
  author = {Aaron van den Oord and Sander Dieleman and Heiga Zen and Karen Simonyan and Oriol Vinyals and Alex Graves and Nal Kalchbrenner and Andrew Senior and Koray Kavukcuoglu},
  title = {WaveNet: A Generative Model for Raw Audio},
  year = {2016},
  archiveprefix = {arXiv},
  eprint = {1609.03499},
  primaryclass = {cs.SD}
}

@article{yu2015multi,
  author = {Yu, Fisher and Koltun, Vladlen},
  title = {Multi-scale context aggregation by dilated convolutions},
  journal = {arXiv preprint arXiv:1511.07122},
  year = {2015}
}

@article{pelletier2019,
  author = {Pelletier, Charlotte and Webb, Geoffrey and Petitjean, Fran{\c c}ois},
  title = {Temporal {Convolutional} {Neural} {Network} for the {Classification} of {Satellite} {Image} {Time} {Series}},
  journal = {Remote Sensing},
  year = {2019},
  volume = {11},
  number = {5},
  pages = {523},
  doi = {10.3390/rs11050523},
  url = {https://www.mdpi.com/2072-4292/11/5/523},
  issn = {2072-4292}
}

@article{zheng1999structural,
  title={Structural basis of DNA recognition by the heterodimeric cell cycle transcription factor E2F--DP},
  author={Zheng, Ning and Fraenkel, Ernest and Pabo, Carl O and Pavletich, Nikola P},
  journal={Genes \& development},
  volume={13},
  number={6},
  pages={666--674},
  year={1999},
  publisher={Cold Spring Harbor Lab}
}
% \bibliography{bibfile}

\end{document}